\documentclass{article}
\usepackage{iclr2027_conference,times}

\usepackage{amsmath,amsfonts,bm}

\def\eqref#1{equation~\ref{#1}}

\def\1{\bm{1}}

\DeclareMathAlphabet{\mathsfit}{\encodingdefault}{\sfdefault}{m}{sl}
\SetMathAlphabet{\mathsfit}{bold}{\encodingdefault}{\sfdefault}{bx}{n}

\usepackage{hyperref}
\usepackage{url}
\usepackage{booktabs}
\usepackage{graphicx}
\usepackage{multirow}
\usepackage{array}
\usepackage{xcolor}
\usepackage{microtype}
\usepackage{enumitem}
\usepackage{amsmath}
\usepackage{caption}
\usepackage{subcaption}
\usepackage{pifont}
\usepackage{makecell}
\usepackage{xspace}
\usepackage{fancyvrb}
\usepackage{fvextra}
\usepackage{needspace}
\usepackage{tcolorbox}
\usepackage{CJKutf8}
\DeclareUnicodeCharacter{2192}{\ensuremath{\rightarrow}}
\tcbuselibrary{breakable,skins}
\usepackage{iftex}
\ifPDFTeX
    \usepackage{CJKutf8}
    \newcommand{\promptlatinfont}{\fontencoding{T1}\fontfamily{pcr}\selectfont}
    \newenvironment{judgechinese}{\begin{CJK*}{UTF8}{gbsn}\fvset{breakanywhere=false}}{\end{CJK*}}
    \newcommand{\judgechinesefont}{}
\else
    \newcommand{\promptlatinfont}{\font\promptmonofont="[lmmono10-regular.otf]" at 8pt\promptmonofont}
    \newenvironment{judgechinese}{="[FandolFang-Regular.otf]" at 8pt\fvset{breakanywhere=true,breakanywheresymbolpre={}}}{}
\fi

\newtcolorbox{prompttemplate}[1]{
    enhanced,breakable,
    title={#1},title after break={#1\ (continued)},
    colback=black!3,colframe=black,
    colbacktitle=black!8,coltitle=black,coltext=black,
    fonttitle=\small\bfseries,fontupper=\fontsize{8}{10}\selectfont\ttfamily,
    arc=1mm,boxrule=0.6pt,boxsep=1mm,
    left=1mm,right=1mm,top=1mm,bottom=1mm,
    before skip=6pt,after skip=8pt
}
\CustomVerbatimCommand{\PromptTemplateInput}{VerbatimInput}{
    fontsize=\fontsize{8}{10}\selectfont,
    formatcom=\color{black}\promptlatinfont,
    breaklines=true,breakautoindent=false,breakindent=0pt,tabsize=4,
    breaksymbolleft={},breaksymbolright={},
    frame=none,xleftmargin=0pt,xrightmargin=0pt
}
\CustomVerbatimEnvironment{PromptTemplateVerbatim}{Verbatim}{
    fontsize=\fontsize{8}{10}\selectfont,
    formatcom=\color{black}\promptlatinfont,
    breaklines=true,breakautoindent=false,breakindent=0pt,tabsize=4,
    breaksymbolleft={},breaksymbolright={},
    frame=none,xleftmargin=0pt,xrightmargin=0pt
}

\newcommand{\benchmark}{\textsc{LifeSearch}\xspace}

\definecolor{yiboblue}{RGB}{45,105,165}

\newcommand{\tablecheck}{\textcolor{green!50!black}{\ding{51}}}
\newcommand{\tablecross}{\textcolor{red!75!black}{\ding{55}}}

\title{\benchmark: Benchmarking Deep Research Agents for Real-World Life and Productivity}

\author{%
\textbf{Yibo Wang, Senjie Jin, Yansong Ning, Xinbing Liang, Haotian Luo,}\\
\textbf{Jiazheng Zhang, Yuxuan HE, Will WU, Clarenceai}\\[4pt]
LLM Department, Tencent
}
\hypersetup{
    pdftitle={LifeSearch: Benchmarking Deep Research Agents for Real-World Life and Productivity},
    pdfauthor={Yibo Wang, Senjie Jin, Yansong Ning, Xinbing Liang, Haotian Luo, Jiazheng Zhang, Yuxuan HE, Will WU, Clarenceai}
}

\iclrfinalcopy
\fancypagestyle{tencenttitle}{%
    \fancyhf{}%
    \fancyhead[L]{\includegraphics[height=20pt]{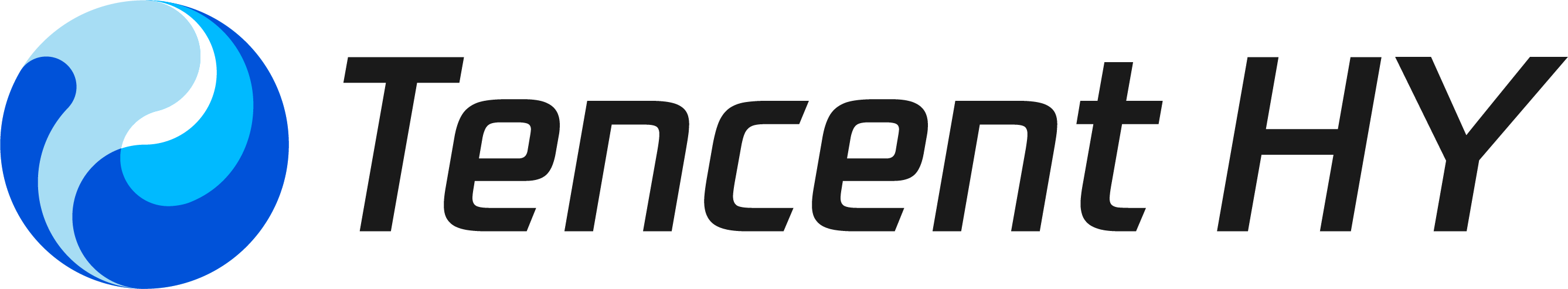}}%
    \fancyfoot[C]{\thepage}%
    \renewcommand{\headrulewidth}{0.4pt}%
    \renewcommand{\footrulewidth}{0pt}%
}
\begin{document}
\maketitle
\pagestyle{plain}
\thispagestyle{tencenttitle}
\begingroup
\renewcommand{\thefootnote}{}
\footnotetext[0]{\textcolor{black}{Correspondence to: yibowang0021@gmail.com}}
\endgroup

\begin{abstract}
Deep research (DR) agents address open-ended information needs through multi-step reasoning, long-horizon web search, and evidence synthesis.
However, existing benchmarks primarily focus on academic research or synthetic tasks, leaving DR agents’ capabilities in \emph{real-world life and professional productivity} settings insufficiently evaluated. 
To bridge this gap, we introduce \benchmark, a challenging benchmark comprising 203 Chinese and English tasks derived from real-user query seeds across 7 concrete life and productivity domains.
Specially, we spend totally more than 4,500 hours of human annotation to construct over 4,000 expert-written, fine-grained rubrics, covering both report-content criteria for coverage and accuracy, as well as concrete criteria specifying values, units, dates, and URLs.
We conduct extensive experiments on 12 advanced DR agents under a unified evaluation harness, systematically analyzing their performance patterns and failure modes.
We find that even the best-performing frontier agent, GPT-5.6 Sol, achieves a score of only 63.4\%, revealing the potential improvement on realistic life and professional DR tasks. 
We compare candidate LLM judges against human rubric labels to select the grading configuration. A human ranking study shows strong alignment between aggregate benchmark scores and human judgments, with Spearman's $\rho = 0.89$ and Kendall's $\tau = 0.81$.
\benchmark provides a rigorous evaluation framework and foundation for advancing DR agents toward real-world life and productivity.

\end{abstract}

\begin{center}
\begin{minipage}{\linewidth}
\centering
\captionsetup{font=small,skip=3pt,hypcap=false}
\captionsetup{type=figure}
\captionsetup[subfigure]{font=small,skip=2pt,justification=centering}
\begin{subfigure}[t]{0.48\linewidth}
\centering
\includegraphics[width=\linewidth]{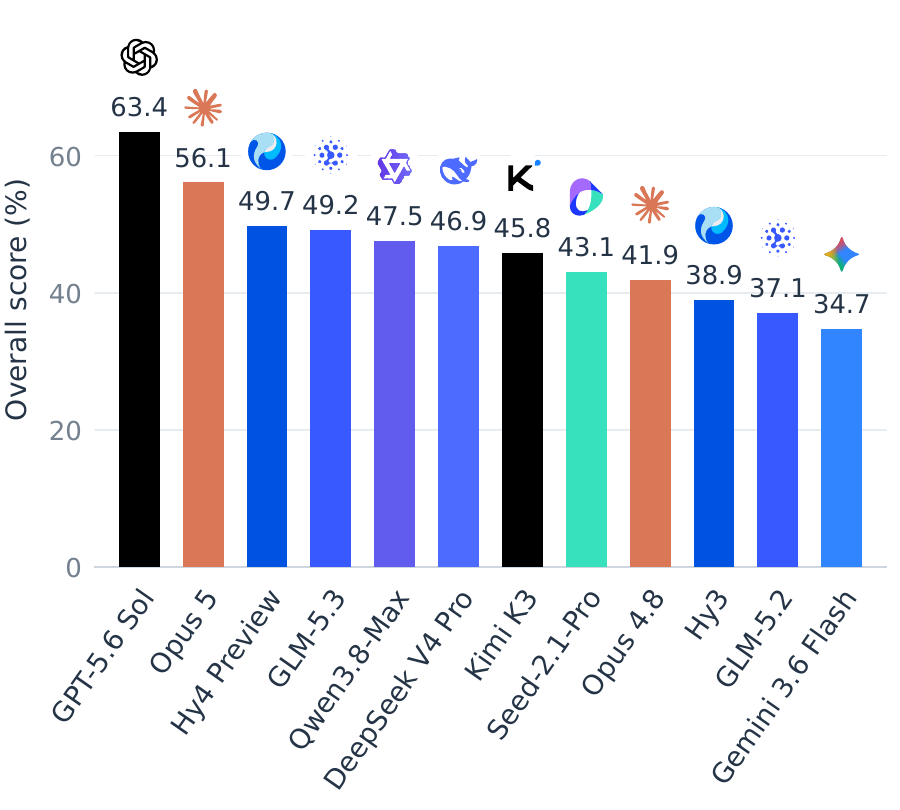}
\caption{Overall performance}
\label{fig:results_teaser_a}
\end{subfigure}\hfill
\begin{subfigure}[t]{0.52\linewidth}
\centering
\includegraphics[width=\linewidth]{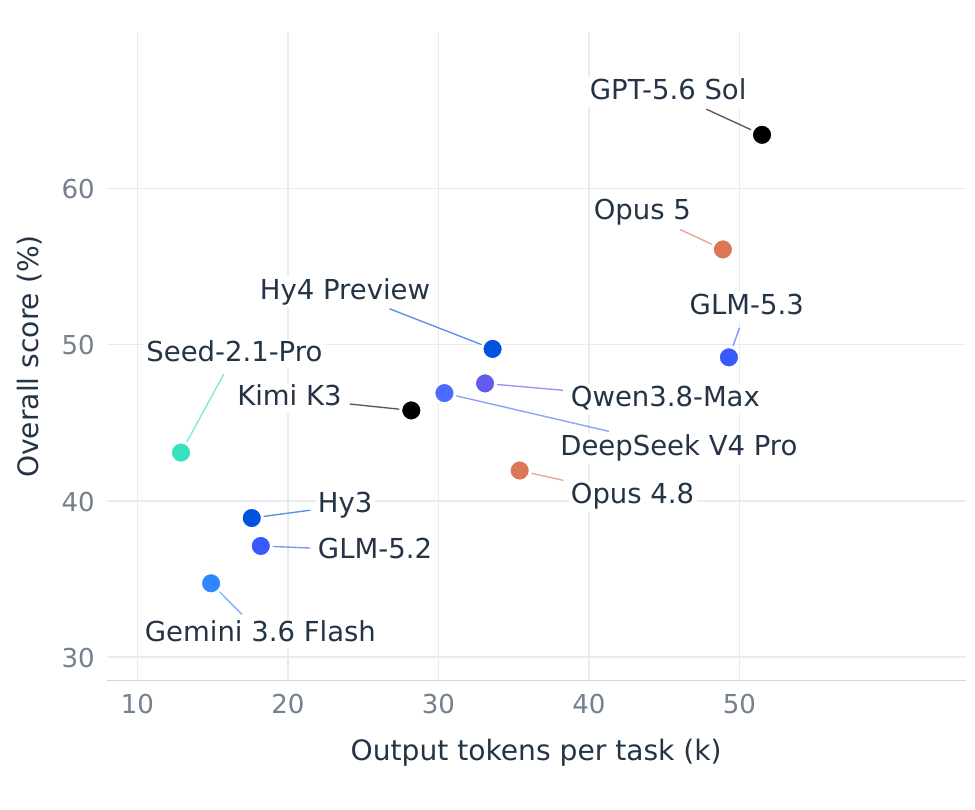}
\caption{Score vs. output tokens}
\label{fig:results_teaser_b}
\end{subfigure}
\caption{\textbf{Performance and generation effort on \benchmark.} Overall scores from Table~\ref{tab:main_results} and mean output tokens per task for the same 12 frontier agents.}
\label{fig:results_teaser}
\end{minipage}
\end{center}

\section{Introduction}
\label{sec:intro}

Deep Research (DR) agents solve open-ended information-seeking problems by iteratively interacting with search engines and web pages, planning multi-step retrieval, and summarizing the collected evidence into a comprehensive report \citep{zhang2025deep,google2025deepresearch, openai2025deepresearch}. 
Building on this agentic search paradigm, DR agents demonstrate great potential across knowledge-intensive domains such as software engineering, biomedicine, and finance \citep{singh2025code,chen2025medbrowsecomp,zhu2025findeepresearch}.
Due to their broad applicability, DR agents attract growing interest from both academia and industry, with increasing efforts devoted to evaluating DR agents.

Many studies evaluate the capability of DR agents.
Early benchmarks primarily rely on tasks with easily verifiable, short-form answers, e.g., BrowseComp~\citep{wei2025browsecomp} poses fact-seeking questions that can be graded by exact string match against a single gold answer.
To better reflect real-world use, subsequent work introduces open-ended, long-horizon tasks and detailed rubrics for evaluating the resulting reports.
For example, DeepResearch Bench~\citep{du2026deepresearch} and ResearchRubrics~\citep{sharma2026researchrubrics} score reports against structured, multi-criterion rubrics, DRACO~\citep{zhong2026draco} targets cross-domain accuracy, completeness, and objectivity, and Mind2Web~2~\citep{gou2025mind2web2} introduces an agent-as-a-judge framework built on tree-structured rubrics.
DeepResearchEval~\citep{wang2026deepresearcheval} introduces an agentic evaluation pipeline that combines adaptive quality assessment with active fact-checking via web search, while WideSearch~\citep{wong2026widesearch} evaluates broad information collection through structured tables.
DR agents are also becoming increasingly capable of addressing practical tasks in everyday life and professional work.
Nevertheless, as summarized in Table~\ref{tab:comparison}, existing benchmarks provide limited coverage of \emph{real-world life and professional productivity} settings, with many evaluations remaining centered on academic research tasks or synthetic tasks.

\begin{table}[t]
\caption{Comparison of \benchmark with representative Deep Research benchmarks.}
\label{tab:comparison}
\centering
\small
\setlength{\tabcolsep}{3pt}
\renewcommand{\arraystretch}{1.15}
\resizebox{\linewidth}{!}{
\begin{tabular}{lccccc}
\toprule
Benchmark & \makecell{Life \& productivity} & \makecell{Real-user query} & \makecell{Expert-written} & \makecell{Chinese \& English} & \makecell{output format} \\
\midrule
BrowseComp~\citep{wei2025browsecomp} & \tablecross & \tablecross & \tablecross & \tablecross & Short answers \\
Mind2Web~2~\citep{gou2025mind2web2} & \tablecross & \tablecross & \tablecross & \tablecross & Lists/tables \\
DeepResearch Bench~\citep{du2026deepresearch} & \tablecross & \tablecross & \tablecross & \tablecheck & Reports \\
WideSearch~\citep{wong2026widesearch} & \tablecheck & \tablecheck & \tablecross & \tablecheck & Tables \\
ResearchRubrics~\citep{sharma2026researchrubrics} & \tablecross & \tablecross & \tablecheck & \tablecross & Report \\
LiveResearchBench~\citep{wang2025liveresearchbench} & \tablecross & \tablecross & \tablecross & \tablecross & Reports \\
DRACO~\citep{zhong2026draco} & \tablecross & \tablecheck & \tablecheck & \tablecross & Reports \\
\midrule
\benchmark (Ours) & \tablecheck & \tablecheck & \tablecheck & \tablecheck & Reports \\
\bottomrule
\end{tabular}
}
\vspace{-6mm}
\end{table}

Motivated by this gap, we present \benchmark, a benchmark for evaluating DR agents on real-world life and professional productivity search. 
We hired crowd workers to write the initial query seeds, which undergo filtering and expert-guided refinement to form the benchmark tasks.
\benchmark spans 7 concrete life and productivity domains, and its construction involves more than 4,500 hours of human labor to produce over 4,000 expert-written, fine-grained rubrics that comprehensively assess both the coverage and the accuracy of a candidate report. 
Specifically, \benchmark comprises 203 tasks: 102 in Chinese and 101 in English.
A substantial fraction of our rubrics are \emph{numeric} and grounded in concrete values, units, or thresholds.
Satisfying these criteria therefore requires retrieving and reporting the precise quantities rather than producing qualitatively plausible statements.
Furthermore, we compare 11 LLM-judge configurations on 402 human-annotated rubric labels and select Gemini 3.1 Pro with low reasoning effort and zero temperature for benchmark grading. A human ranking study shows that the overall rankings produced by our evaluation closely agree with human rankings, yielding Spearman's $\rho=0.89$ and Kendall's $\tau=0.81$.

We evaluate 12 frontier DR agents under a shared harness (Figure~\ref{fig:results_teaser}), yielding three main findings. \textbf{i) Substantial headroom remains:} Even top-performing GPT-5.6 Sol reaches only 63.4\%, with all systems falling below 70\% across domains.
\textbf{ii) Quality transcends volume:} Higher scores correlate with longer reports, yet volume alone is non-deterministic: Claude Opus 5 and GLM-5.3 have comparable token counts but differ by 6.9 percentage points, highlighting the centrality of evidence synthesis.
\textbf{iii) Common failure modes:} Autonomous runs can fail through premature clarification, use of evidence outside the requested time window, and omissions of specific task requirements.

In summary, our contributions are:
\begin{itemize}[leftmargin=*,nosep]
    \item \textbf{The first benchmark centered on real-world life and professional productivity DR evaluation.} \benchmark shifts the evaluation lens of DR agents toward realistic life and work search, and reveals a concrete capability bottleneck of current frontier models in this setting.
    \item \textbf{\benchmark: a large-scale, expert-annotated bilingual benchmark.} We construct and open-source \benchmark, spanning 7 life and productivity domains and comprising more than 4,000 expert-written, fine-grained rubrics produced through over 4,500 hours of human annotation labor, to comprehensively evaluate both the coverage and the accuracy of DR agent reports.
    \item \textbf{A comprehensive empirical study.} We carefully examine agreement with human judgments at both the rubric and overall-ranking levels, systematically benchmark 12 current DR agents under a unified harness, providing actionable insight for the DR research community.
\end{itemize}

\section{Related Work}

\textbf{Deep Research Agent.}
Deep Research (DR) agents extend earlier web-browsing assistants into autonomous systems that interleave multi-step reasoning with iterative search and synthesize the retrieved evidence into a research report, e.g., GAIA~\citep{mialon2023gaia}, WebArena~\citep{zhou2024webarena}, and AssistantBench~\citep{yao2024assistantbench}.
Early work such as Search-R1~\citep{jin2025search} and Kimi-Researcher~\citep{kimiResearcher} trains LLMs to interleave reasoning with search-engine calls, enabling emerging agentic research capabilities.
Building on this, subsequent studies such as Tongyi DeepResearch~\citep{team2025tongyi}, together with harder benchmarks such as BrowseComp~\citep{wei2025browsecomp}, push DR agents toward long-horizon and more difficult information-seeking behavior.
DR agents show strong potential across knowledge-intensive domains, including biomedicine and finance~\citep{chen2025medbrowsecomp,zhu2025findeepresearch}.

\textbf{Deep Research Agent Evaluation.}
As DR agents increasingly produce open-ended, long-form reports, evaluating their outputs becomes substantially more challenging than verifying short, exact answers. 
Subsequent work further decomposes each task into a structured set of criteria: DeepResearch Bench~\citep{du2026deepresearch} and ResearchRubrics~\citep{sharma2026researchrubrics} grade reports against multi-criterion rubrics, ReportBench~\citep{li2025reportbench} evaluates academic survey reports, DRACO~\citep{zhong2026draco} targets cross-domain accuracy, completeness, and objectivity, and Mind2Web~2~\citep{gou2025mind2web2} introduces an agent-as-a-judge framework over tree-structured rubrics.
More recent work further automates this pipeline: DeepResearchEval~\citep{wang2026deepresearcheval} combines adaptive quality assessment with active fact-checking, while LiveResearchBench~\citep{wang2025liveresearchbench} and WideSearch~\citep{wong2026widesearch} extend rubric-style grading to dynamic, user-centric reports and structured-table deliverables, respectively. 
However, most of these focus on academic research or synthetic tasks, leaving DR agents' capabilities in \emph{real-world life and professional productivity} settings underexplored. Their rubrics also lack the numerical grounding, and independent verifiability needed to assess whether a report is truly usable as a real-world life or work deliverable.

\begin{figure}[t]
    \centering
    \includegraphics[width=\linewidth]{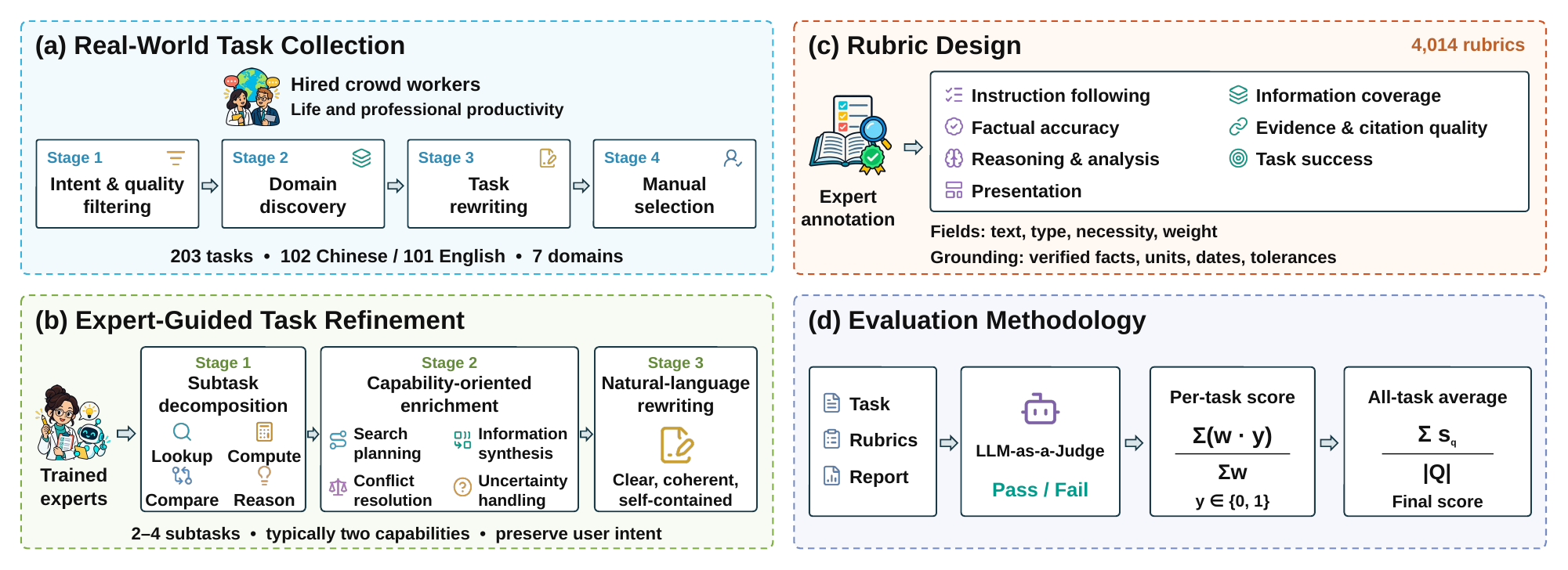}
    \caption{Overview of \benchmark. (a) Collect and filter query seeds written by hired crowd workers. (b) Refine queries into research tasks with expert guidance. (c) Construct fine-grained rubrics. (d) Evaluate reports with an LLM-as-a-Judge and aggregate rubric judgments into scores.}
    \label{fig:categories}
\vspace{-4mm}
\end{figure}

\section{Overview of \benchmark}
\label{sec:benchmark_overview}

\begin{figure}[t]
    \centering
    \captionsetup[subfigure]{font=small,labelfont=normalfont,justification=centering,skip=3pt}
    \begin{subfigure}[t]{0.48\linewidth}
        \centering
        \includegraphics[width=\linewidth]{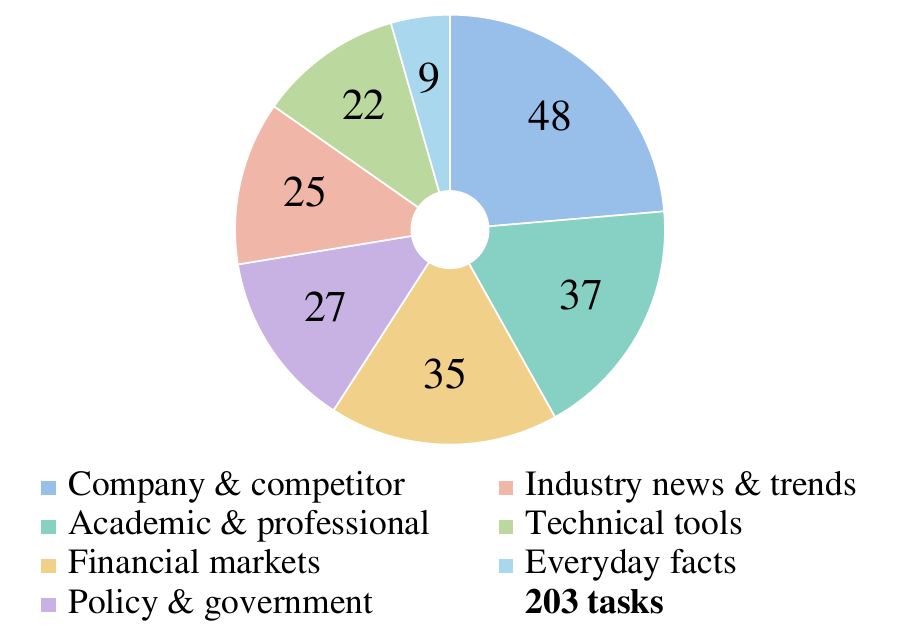}
        \caption{Task distribution}
        \label{fig:task_distribution}
    \end{subfigure}\hfill
    \begin{subfigure}[t]{0.48\linewidth}
        \centering
        \includegraphics[width=\linewidth]{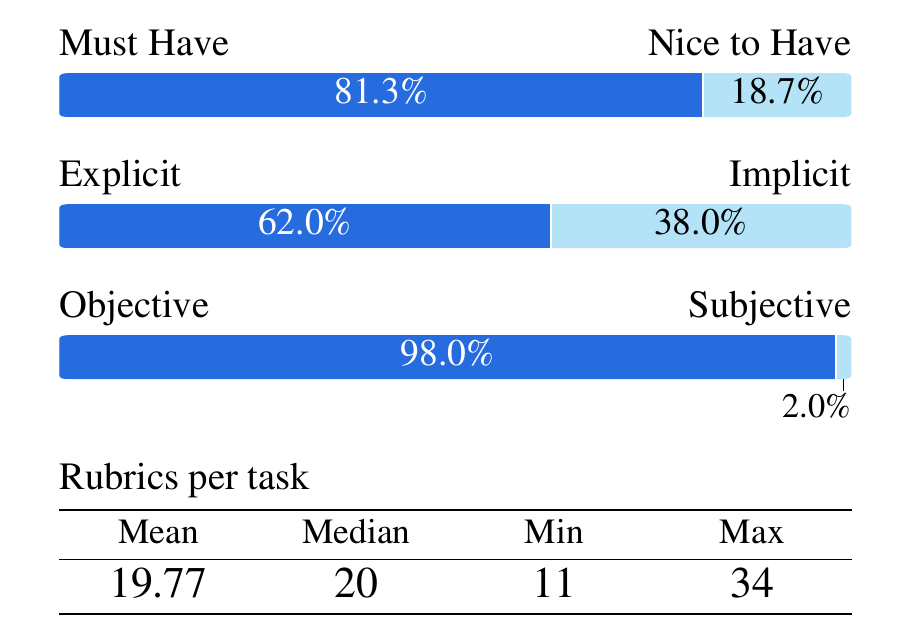}
        \caption[Rubric composition]{Rubric composition.}
        \label{fig:rubric_statistics}
    \end{subfigure}
    \caption{Dataset composition of \benchmark. (a) Task counts across seven domains. (b) Annotation attributes across 4,014 rubrics and summary statistics of rubrics per task. Constructing the benchmark required more than 4,500 hours of human annotation.}
    \label{fig:dataset_profile}
\vspace{-7mm}
\end{figure}

Figure~\ref{fig:categories} outlines the construction and evaluation pipeline, while Figure~\ref{fig:dataset_profile} summarizes the task distribution and rubric statistics. \benchmark comprises 203 tasks (102 in Chinese and 101 in English) across seven real-world personal and professional productivity domains. The tasks are developed from query seeds written by hired crowd workers and refined by experts to express complex, pragmatic information-seeking needs. To rigorously evaluate generated reports, trained expert annotators develop 4,014 fine-grained rubrics assessing both requirement coverage and factual accuracy. Crucially, many rubrics are anchored to concrete numerical values, units, dates, or thresholds, facilitating verifiable, criterion-level evaluation.

\subsection{Data Collection and Task Domains}
\label{sec:data_collection}
To target real-world life and professional productivity scenarios, \benchmark requires both authentic task provenance and broad coverage of user needs in its construction. We begin with requests written by hired crowd workers. These requests span diverse workflows---including file operations, software engineering, and deep research---providing a heterogeneous pool of realistic user needs.

We construct the data collection pipeline through the following stages:

\textbf{Stage 1: Intent and quality filtering.} We use an LLM to assess search relevance, query quality, and dependence on implicit recent events. We filter out requests unrelated to information seeking, queries with unclear intent or insufficient context, and requests that presuppose an unspecified recent event or state change. General time-sensitive searches remain eligible. Appendix~\ref{app:filter_prompt} provides the  prompt.

\textbf{Stage 2: Domain discovery.} We use an LLM to classify the retained queries by their core purpose into seven productivity domains: company and competitor research, academic and professional literature, financial market analysis, policy and government data, industry news and trend aggregation, technical tools and developer resources, and everyday facts. Appendix~\ref{app:domain_prompt} provides the template.

\textbf{Stage 3: Task rewriting.} Candidate queries may combine information seeking with operational actions such as generating local files or interactive HTML visualizations. We use an LLM to remove these operations and local-file dependencies with minimal changes, preserving the research intent, source constraints, temporal expressions, and content-organization requirements. Appendix~\ref{app:rewrite_prompt} provides the rewriting template.

\textbf{Stage 4: Manual selection.} Experienced AI researchers perform a final screening to retain tasks that are sufficiently complex, diverse, clear, and feasible to evaluate.

Figure~\ref{fig:dataset_profile} summarizes the distribution of the retained tasks across the seven domains. The resulting mix spans both specialized research for professional decision-making and practical information needs arising in everyday life, rather than concentrating on a single subject area.

\subsection{Expert-Guided Task Refinement}
\label{sec:task_refinement}
We invite experienced annotation experts to refine the query seeds retained in Section~\ref{sec:data_collection} into self-contained research tasks while preserving the original query intent. The same experts also participate in rubric annotation (Section~\ref{sec:rubric_design}). The refinement process comprises three stages.

\textbf{Stage 1: Subtask decomposition.}
Each seed is decomposed into a small set of complementary subtasks, typically two to four, each with a concise description and one of four types: \textit{Lookup} for retrieving entities, facts, or values; \textit{Compute} for deriving quantitative indicators; \textit{Compare} for examining differences across entities or time periods; and \textit{Reason} for drawing conclusions and assessing implications. This decomposition makes the information needs explicit and helps identify missing or redundant dimensions.

\textbf{Stage 2: Capability-oriented enrichment.}
Experts refine the subtasks around a small set of relevant capabilities, typically two chosen from \textit{Search Planning} (SP), \textit{Information Synthesis} (IS), \textit{Conflict Resolution} (CR), and \textit{Uncertainty Handling} (UH). SP emphasizes comprehensive coverage of parallel aspects; IS requires integrating evidence from multiple sources; CR introduces source comparison and authority constraints; and UH requires acknowledging missing evidence, providing justified estimates, or stating limitations. Redundant subtasks are merged or removed, and additional requirements are introduced only when needed to probe the selected capabilities. The task thus gains depth while remaining focused on the user's original goal.

\textbf{Stage 3: Natural-language rewriting.}
Finally, experts consolidate the refined subtasks into a coherent query grounded in a concrete life or work scenario. The wording preserves the user's perspective and naturally incorporates the relevant entities, source requirements, and temporal scope where needed. Final revisions improve clarity, remove repetitive or formulaic phrasing, and ensure that the intended capability requirements remain explicit in the resulting request.

\subsection{Rubric Design}
\label{sec:rubric_design}
For each refined task $q$, the experts involved in task refinement construct a rubric set $R_q$ describing the requirements of a satisfactory report. Finance experts also participate in annotating rubrics for financial tasks. We use \emph{rubric} to denote a single scoring requirement that supports an independent Pass/Fail judgment. The benchmark contains 4,014 rubrics, averaging 19.8 per task. Table~\ref{tab:rubric_design} summarizes the assessment dimensions and annotation fields; detailed writing requirements and illustrative examples are provided in Appendix~\ref{app:rubric_design}. Appendix~\ref{app:task_rubric_examples} presents Chinese and English tasks with selected rubrics and their annotation attributes.

\textbf{Assessment dimensions.}
Rubric construction covers seven complementary dimensions: instruction following, information coverage, factual accuracy, evidence and citation quality, reasoning and analysis, task success, and presentation.

\textbf{Rubric representation.}
Each rubric records its text (stored as \texttt{criterion}), the acceptance statement; \texttt{type}, distinguishing \textsc{Objective} from \textsc{Subjective} judgments; \texttt{necessity}, distinguishing \textsc{Explicit} requests from \textsc{Implicit} but reasonable requirements; \texttt{priority}, retaining the qualitative \textsc{Must Have} / \textsc{Nice to Have} label supplied to the judge; and \texttt{weight}, an annotator-assigned measure of the rubric's relative importance within the task. Numerical aggregation uses \texttt{weight}, rather than deriving weights from the priority label. The guidelines prioritize objective rubrics and reserve unavoidable subjective preferences for optional aspects.

\textbf{Weight assignment.}
Annotators assign each rubric a weight according to its relative importance for fulfilling the user's request. We record weights as floating-point values in $[0,1]$ with one decimal place, yielding increments of $0.1$. Higher weights indicate more important requirements. These numerical weights are used directly for score aggregation, as detailed in Section~\ref{sec:score_aggregation}.

\textbf{Writing and grounding requirements.}
Rubrics are self-contained, non-redundant, and operationally precise. Experts define target entities, factual targets, units, temporal bounds, and tolerances so that judgments require no additional web browsing. Expert annotators specify and verify factual requirements using authoritative sources. Approximately 40\% of the rubrics explicitly specify numerical values, specific events, or named documents, which provide concrete targets for verification.

For open-ended tasks such as recommendations, rubrics evaluate comparative depth and constraint satisfaction rather than a single rigid response, thereby permitting valid alternatives. Rubrics tied to specific examples remain optional unless the prompt explicitly requires them.

\begin{table}[t]
\caption{Rubric assessment dimensions and annotation fields.}
\label{tab:rubric_design}
\centering
\small
\setlength{\tabcolsep}{5pt}
\renewcommand{\arraystretch}{1.12}
\begin{tabular}{@{}>{\raggedright\arraybackslash}p{0.30\linewidth}>{\raggedright\arraybackslash}p{\dimexpr0.70\linewidth-10pt\relax}@{}}
\toprule
\textbf{Dimension / field} & \textbf{Definition} \\
\midrule
\multicolumn{2}{@{}l}{\textbf{A. Assessment dimensions}} \\
\midrule
Instruction following & Fulfillment of explicit user requests. \\
Information coverage & Coverage of the requested entities, attributes, and scope. \\
Factual accuracy & Correctness of concrete facts and numerical values. \\
Evidence and citation quality & Required source identity, attribution, and temporal context. \\
Reasoning and analysis & Logical connections between evidence and conclusions. \\
Task success & Resolution of the user's decision or question. \\
Presentation & Compliance with requested structure and format. \\
\midrule
\multicolumn{2}{@{}l}{\textbf{B. Annotation fields}} \\
\midrule
\texttt{criterion} & A self-contained acceptance statement for a Pass/Fail judgment. \\
\texttt{type} & \textsc{Objective} / \textsc{Subjective}: verifiable facts or requirements versus judgments involving subjective preferences. \\
\texttt{necessity} & \textsc{Explicit} / \textsc{Implicit}: requirements stated in the query versus those reasonably implied by the task. \\
\texttt{priority} & \textsc{Must Have} / \textsc{Nice to Have}: qualitative annotation retained in the judge prompt, distinct from the numerical weight. \\
\texttt{weight} & Annotator-assigned relative importance within the task: a floating-point value in $[0,1]$, recorded to one decimal place. \\
\bottomrule
\end{tabular}
\vspace{-3mm}
\end{table}

\textbf{Quality control.}
We implement annotation quality control across two stages:
(i) an iterative cycle of annotation, inspection, and revision to refine the rubrics; 
(ii) outlier screening, where low-scoring tasks are checked for flaws, while overly easy tasks are tightened.

\subsection{Evaluation Methodology}
\label{sec:evaluation_methodology}
\label{sec:score_aggregation}
We adopt a rubric-based LLM-as-a-Judge approach to assess each final report. The resulting rubric-level judgments are aggregated into a score.

Given a task, its rubric set, and a candidate report, an LLM judge assigns a binary Pass/Fail label to each rubric. Judgments follow the rubric's acceptance conditions, including factual targets, source requirements, and temporal scope. Semantically equivalent wording and numerical approximations are accepted where permitted by the rubric. Let $R_q$ denote the rubric set for task $q$, and let $y_{q,r}\in\{0,1\}$ indicate whether the candidate report for task $q$ satisfies rubric $r$.

For each rubric $r$ in task $q$, we use its annotator-assigned weight $w_{q,r}$, which reflects relative importance within the task. We first normalize the weighted satisfaction score within each task and then average these scores uniformly across the task set $Q$:
\begin{equation}
s_q
=
\frac{\sum_{r\in R_q}w_{q,r}y_{q,r}}
     {\sum_{r\in R_q}w_{q,r}},
\qquad
\mathrm{score}
=
\frac{1}{|Q|}
\sum_{q\in Q}s_q.
\label{eq:score}
\end{equation}
The denominator sums the weights of all rubrics in the task (including unmet ones), requiring a positive total weight for well-defined normalization; zero-weight rubrics are omitted. Averaging $s_q$ uniformly across tasks assigns equal importance to each task regardless of rubric count or scale, measuring weighted requirement satisfaction under annotated importance weights.

\subsection{Judge Model Selection in DR Setting}
\label{sec:judge_selection}
For judge selection in the DR rubric setting, we annotate one anonymized model response for each of 20 tasks (10 Chinese, 10 English), covering 402 rubrics. For each report, two to three domain experts independently assign binary Pass/Fail labels, reaching a substantial inter-annotator agreement (mean pairwise Cohen's $\kappa=0.715$). All 65 initial disagreements are subsequently resolved through discussion to form the reference labels used to compare judge configurations.

Benchmarking against these reference labels, we evaluate four frontier LLMs across 11 configurations of reasoning effort and temperature (Table~\ref{tab:judge_selection}). Each configuration is run three times on identical reports. We assess performance using three metrics: mean accuracy against human annotations, self-consistency (the proportion of rubrics yielding identical decisions across all three runs), and majority-voted accuracy at the rubric level.

\begin{table}[t]
\centering
\caption{Comparison of 11 judge configurations on 402 human-labeled rubrics for judge selection, sorted by mean accuracy over three runs. Default denotes the model's default setting. Bold numbers mark the best result in each metric.}
\label{tab:judge_selection}
\small
\setlength{\tabcolsep}{4pt}
\renewcommand{\arraystretch}{1.1}
\begin{tabular}{@{}lccccc@{}}
\toprule
Judge model & Reasoning & Temp. & \makecell{Mean accuracy (\%)} & \makecell{Self-consistency (\%)} & \makecell{Majority vote (\%)} \\
\midrule
Gemini 3.1 Pro & Low & 0 & \textbf{84.25} & 98.8 & \textbf{84.33} \\
Gemini 3.1 Pro & High & Default & 84.00 & 94.5 & \textbf{84.33} \\
Gemini 3.1 Pro & High & 0 & 83.50 & 98.0 & 83.33 \\
Gemini 3.5 Flash & High & Default & 82.42 & 94.8 & 82.34 \\
DeepSeek V4.1 Flash & High & 0 & 81.43 & 94.8 & 81.34 \\
DeepSeek V4.1 Flash & High & Default & 81.43 & 91.3 & 81.84 \\
DeepSeek V4.1 Flash & Low & 0 & 80.10 & 94.0 & 80.10 \\
GPT-5.5 & High & Default & 79.35 & 95.3 & 79.10 \\
DeepSeek V4.1 Flash & Off & 0 & 78.61 & 97.0 & 78.61 \\
GPT-5.5 & Off & 0 & 75.95 & 98.8 & 76.37 \\
Gemini 3.5 Flash & Off & 0 & 75.62 & \textbf{99.0} & 75.62 \\
\bottomrule
\end{tabular}
\vspace{-4mm}
\end{table}

\textbf{Temperature and stability.}
Setting temperature to zero improves self-consistency—e.g., rising from 94.5\% to 98.0\% for Gemini 3.1 Pro and 91.3\% to 94.8\% for DeepSeek V4.1 Flash at high reasoning effort. However, stability does not guarantee accuracy: Gemini 3.5 Flash without reasoning achieves 99.0\% self-consistency but only 75.62\% mean accuracy.

\textbf{Limited gains from majority voting.}
Majority voting yields marginal improvements over mean accuracy ($\le$0.42 percentage points across all configurations), merely shifting accuracy from 84.25\% to 84.33\% for Gemini 3.1 Pro (low reasoning, zero temperature). We therefore adopt a single judge pass per report for benchmark evaluation.

Based on this comparison, we select Gemini 3.1 Pro with low reasoning effort and zero temperature, which achieves the highest mean accuracy (84.25\%) among the tested configurations on these 402 labels. We use this configuration for benchmark grading. This model choice aligns with findings by \citet{peng2026can} in DR settings.

\section{Experiments}
\label{sec:experiments}
We evaluate frontier Deep Research agents on \benchmark to assess their ability to complete real-world tasks and examine agreement between aggregate scores and human report rankings. We first describe the shared execution harness and experimental setup (Section~\ref{sec:experimental_setup}), then present the main comparative results (Section~\ref{sec:main_results}).
We next examine performance patterns in evidence collection and report synthesis (Section~\ref{sec:performance_patterns}), followed by an analysis of recurring failures (Section~\ref{sec:failure_analysis}). Finally, we examine the alignment between aggregate scores and report-level human preferences (Section~\ref{sec:judge_validation}).

\subsection{Experimental Setup}
\label{sec:experimental_setup}

We evaluate 12 frontier DR agents on all 203 bilingual tasks: GPT-5.6 Sol, Claude Opus 5, Hy4 Preview, GLM-5.3, Qwen3.8-Max, DeepSeek-V4 Pro, Kimi-K3, Doubao-Seed-2.1-Pro, Claude Opus 4.8, Hy3, GLM-5.2, and Gemini-3.6-Flash. We use each model's default configuration, setting reasoning effort to the maximum supported level where configurable. All systems operate within a shared ReAct-style~\citep{DBLP:conf/iclr/YaoZYDSN023} harness that interleaves decisions, tool execution, and feedback from the resulting observations. Agents have access to \textit{web search}, \textit{web fetch}, and \textit{code execution}. Each iteration corresponds to a logical model decision step that may issue multiple tool calls or produce the final answer.
A unified system prompt in Appendix~\ref{app:agent_system_prompt} instructs agents to complete the DR task. We use Serper\footnote{\url{https://serper.dev/}} as the web-search backend and Jina Reader\footnote{\url{https://jina.ai/reader/}} to retrieve web pages.

We evaluate final reports using the score defined in Section~\ref{sec:score_aggregation}, applying a single grading pass over each report's rubrics with Gemini 3.1 Pro~\citep{googledeepmind2026gemini31pro}, as selected in Section~\ref{sec:judge_selection}. To account for variability across DR agent trajectories, we conduct three independent runs per task, and report the mean and sample standard deviation of the run-level benchmark scores. We also record iteration counts and model-generated output tokens. 

\begin{table}[t!]
\centering
\caption{
Performance and resource usage on \benchmark. Scores are reported as mean $\pm$ sample standard deviation across three independent agent runs. Resource statistics are per-task means; output tokens include generation throughout execution, not only the final report. Domain abbreviations: Comp., company and competitor research; Acad., academic and professional literature; Fin., financial market analysis; Pol., policy and government data; Ind., industry news and trend aggregation; Tech., technical tools and developer resources; Everyd., everyday facts.
}
\label{tab:main_results}

\begingroup
\fontsize{8}{9.5}\selectfont
\setlength{\tabcolsep}{2pt}
\renewcommand{\arraystretch}{1.12}

\begin{tabular*}{\linewidth}{@{\extracolsep{\fill}}l*{10}{c}@{}}
\toprule
& \multicolumn{7}{c}{Domain scores (\%) $\uparrow$}
& \textbf{Overall}
& \multicolumn{2}{c}{\textbf{Resource usage}} \\
\cmidrule(lr){2-8}
\cmidrule(lr){9-9}
\cmidrule(l){10-11}
\textbf{Model}
& Comp. & Acad. & Fin. & Pol. & Ind. & Tech. & Everyd.
& Score (\%) $\uparrow$
& Iter.
& \makecell{Output\\tokens (k)} \\
\midrule
GPT-5.6 Sol
    & 62.9 & 67.8 & 60.7 & 65.6 & 59.5 & 65.6 & 58.3
    & $\mathbf{63.44}\;{\scriptstyle\pm\,0.27}$
    & 86.2 & 51.5 \\
Claude Opus 5
    & 52.4 & 60.7 & 55.7 & 58.2 & 52.6 & 59.7 & 53.7
    & $56.10\;{\scriptstyle\pm\,0.28}$
    & 39.0 & 48.9 \\
Hy4 Preview
    & 46.7 & 51.4 & 48.1 & 52.9 & 51.0 & 51.9 & 47.1
    & $49.73\;{\scriptstyle\pm\,0.86}$
    & 40.0 & 33.6 \\
GLM-5.3
    & 47.8 & 49.0 & 47.5 & 52.4 & 48.5 & 50.9 & 51.9
    & $49.19\;{\scriptstyle\pm\,1.20}$
    & 22.3 & 49.3 \\
Qwen3.8-Max
    & 48.2 & 47.3 & 46.9 & 50.4 & 47.1 & 45.4 & 45.1
    & $47.52\;{\scriptstyle\pm\,1.64}$
    & 25.0 & 33.1 \\
DeepSeek-V4 Pro
    & 45.7 & 44.8 & 47.4 & 48.7 & 49.8 & 46.1 & 48.8
    & $46.90\;{\scriptstyle\pm\,1.18}$
    & 21.1 & 30.4 \\
Kimi-K3
    & 44.5 & 44.1 & 42.3 & 49.1 & 49.4 & 46.4 & 51.6
    & $45.79\;{\scriptstyle\pm\,0.76}$
    & 18.7 & 28.2 \\
Doubao-Seed-2.1-Pro
    & 44.4 & 45.2 & 37.2 & 44.1 & 41.8 & 44.1 & 48.4
    & $43.09\;{\scriptstyle\pm\,0.49}$
    & 23.7 & 12.9 \\
Claude Opus 4.8
    & 40.0 & 43.3 & 38.4 & 45.9 & 42.8 & 43.3 & 43.1
    & $41.94\;{\scriptstyle\pm\,0.66}$
    & 12.1 & 35.4 \\
Hy3
    & 40.2 & 38.0 & 35.3 & 41.5 & 40.4 & 37.5 & 41.1
    & $38.90\;{\scriptstyle\pm\,0.48}$
    & 31.0 & 17.6 \\
GLM-5.2
    & 37.1 & 37.6 & 32.4 & 39.5 & 37.2 & 39.8 & 39.5
    & $37.11\;{\scriptstyle\pm\,1.16}$
    & 11.0 & 18.2 \\
Gemini-3.6-Flash
    & 32.8 & 36.0 & 33.2 & 35.1 & 36.5 & 32.7 & 44.5
    & $34.72\;{\scriptstyle\pm\,0.16}$
    & 17.8 & 14.9 \\
\bottomrule
\end{tabular*}

\endgroup
\vspace{-4mm}
\end{table}

\subsection{Main Results}
\label{sec:main_results}

\textbf{Substantial headroom remains.} Table~\ref{tab:main_results} summarizes overall scores, domain performance, and resource usage. GPT-5.6 Sol leads with 63.4\%, followed by Claude Opus 5 at 56.1\%; all other systems score below 50\%. Even the strongest agent leaves 36.6\% of the normalized requirement weight unsatisfied. Its three-run mean score also falls below 50\% on 41 of 203 tasks (20.2\%), indicating that its overall lead does not translate into reliable completion of every research request.

\textbf{Domain results reveal distinct strengths.} GPT-5.6 Sol and Claude Opus 5 rank first and second in all seven domains, but no system exceeds 70\% in any domain. Other systems exhibit more varied strengths: GLM-5.3 outperforms Hy4 Preview on everyday facts (51.9\% versus 47.1\%), while Hy4 Preview leads on academic and professional literature (51.4\% versus 49.0\%). These differences are descriptive, especially for the nine-task everyday-facts subset.

\textbf{Output volume does not determine quality.} Figure~\ref{fig:results_teaser_b} relates scores to model-generated output tokens across the full agent run. GPT-5.6 Sol has both the highest score and the largest output volume (51.5k tokens per task). However, Claude Opus 5 and GLM-5.3 generate similar volumes (48.9k and 49.3k), yet their scores differ by 6.9 percentage points. Hy4 Preview reaches a comparable score to GLM-5.3 (49.7\% versus 49.2\%) with only 33.6k tokens. Thus, greater generation volume alone does not explain performance.

\subsection{Performance Patterns}
\label{sec:performance_patterns}
\benchmark rewards correct, comprehensive answers supported by key evidence and traceable sources. Meeting these requirements involves two complementary capabilities: collecting relevant information through search and synthesizing it into a report that explicitly addresses the user's needs. The observed performance patterns help explain why resource usage alone is insufficient to characterize research quality.

\textbf{Evidence collection.} Repeated searches and source inspection allow agents to cover complementary aspects of a task, verify facts, and resolve gaps or conflicting evidence. Across the 12 systems, mean interaction count correlates positively with overall score (Spearman's $\rho=0.76$). GPT-5.6 Sol averages 86.2 iterations and scores 63.4\%, whereas GLM-5.2 averages 11.0 iterations and scores 37.1\%. This pattern is consistent with the value of sustained evidence collection. However, iterations include other tool actions, and comparisons across models do not isolate the causal effect of additional searches.

\textbf{Evidence synthesis.} Retrieved information contributes to the score only when the final report presents the facts and supporting evidence needed to satisfy the rubrics. Final-report length, measured in characters rather than tokens generated throughout execution (Appendix~\ref{app:response_length}), correlates positively with score (Spearman's $\rho=0.60$). Yet Doubao-Seed-2.1-Pro and Gemini-3.6-Flash produce 12,089 and 10,256 characters per report while scoring 43.1\% and 34.7\%, respectively; Hy4 Preview reaches 49.7\% with 9,959 characters. Thus, longer reports can accommodate broader coverage, but their value depends on the relevance, correctness, and completeness of the information they contain.

\subsection{Failure Analysis}
\label{sec:failure_analysis}
Our analysis identifies three recurring failure modes: premature termination, temporal scope violations, and incomplete factual coverage. They affect whether an agent produces a final report, whether its evidence is valid for the requested period, and whether the report satisfies the task's requirements.

\textbf{Premature clarification.} Doubao-Seed-2.1-Pro and GPT-5.6 Sol sometimes ask follow-up questions instead of proceeding with the research task. Because the autonomous harness provides no user replies, these requests can terminate the loop before a final report is produced. The agent consequently leaves the task unresolved and provides no research deliverable for evaluation.

\textbf{Temporal scope violations.} Agents sometimes default to current web data even when the query specifies an earlier cutoff, a behavior particularly noticeable in Hy4 Preview. Retrieved information can be accurate at execution time yet invalid for the requested period. Using updated figures or later events without checking their dates can therefore make an otherwise relevant report inconsistent with the task's temporal scope.

\textbf{Missing required details.} With around 20 rubrics per task, a fluent summary can address the broad topic while omitting requested entities, numerical values, dates, sources, or edge cases. These omissions leave individual requirements unmet even when the overall narrative appears coherent. 

\subsection{Human Agreement}
\label{sec:judge_validation}

\noindent
\begin{minipage}[t]{0.55\linewidth}
\vspace{0pt}
We conduct a human ranking study on 30 tasks (15 per language) to examine agreement at the report level. Annotators rank anonymized reports presented in randomized order according to factual accuracy, completeness, and formatting quality. This assessment examines whether aggregate rubric scores capture the relative quality of complete research deliverables.

Table~\ref{tab:human_machine_ranks} presents a seven-system study, comprising 210 reports. Within this subset, rubric-based and human rankings are strongly correlated (Spearman's $\rho=0.89$, Kendall's $\tau=0.81$), with identical ordering of the top four systems. These results support the alignment of our aggregate rubric-based scores with human assessments of overall report quality.
\end{minipage}\hfill%
\begin{minipage}[t]{0.40\linewidth}
\vspace{0pt}
\centering
\captionsetup{font=small,skip=5pt,hypcap=false}
\captionof{table}{Ranks among the seven displayed models. $\Delta$ shows the change from Ours to Human: $\uparrow$ better, $\downarrow$ worse.}
\label{tab:human_machine_ranks}
\fontsize{8}{9.5}\selectfont
\setlength{\tabcolsep}{3pt}
\renewcommand{\arraystretch}{1.15}
\begin{tabular*}{\linewidth}{@{\extracolsep{\fill}}lccc@{}}
\toprule
\textbf{Model}
& \makecell{\textbf{Human}\\\textbf{rank}}
& \textbf{Ours}
& $\boldsymbol{\Delta}$ \\
\midrule
GPT-5.6 Sol         & 1 & 1 & --- \\
Claude Opus 5      & 2 & 2 & --- \\
Doubao-Seed-2.1-Pro & 3 & 3 & --- \\
Kimi-K3            & 4 & 4 & --- \\
Hy3                & 7 & 5 & $\downarrow 2$ \\
Claude Opus 4.8    & 5 & 6 & $\uparrow 1$ \\
GLM-5.2            & 6 & 7 & $\uparrow 1$ \\
\bottomrule
\end{tabular*}
\vspace{-4mm}
\end{minipage}
\par




\section{Conclusion}
We present \benchmark, a benchmark for evaluating deep research agents on real-world life and productivity tasks. Spanning seven domains, it comprises 203 bilingual tasks (Chinese and English) with 4,014 expert-annotated rubrics assessing requirement coverage and factuality. Across 12 frontier agents evaluated in a unified harness, the top system scores only 63.4\%, with increased output volume failing to consistently ensure higher quality. Our grading protocol is validated against human rubric labels and holistic report rankings. We hope \benchmark supports the development of more reliable and effective DR agents for everyday life and professional work.


\subsection*{AI use statement}
We use ChatGPT solely for language refinement of the manuscript's text. All conceptual content, experimental design, analysis, and conclusions are developed entirely by the authors. We carefully review the AI-assisted edits to ensure that the meaning and technical accuracy of the original text are fully preserved.

\subsection*{Ethics statement}
We hired crowd workers to write the initial query seeds for \benchmark. All candidate queries undergo privacy screening and anonymization before task construction. Expert annotators responsible for task refinement and rubric annotation are full-time employees and performed annotation as part of their compensated work.

\subsection*{Reproducibility statement}
Section~\ref{sec:benchmark_overview} details the data collection, expert-guided task refinement, rubric construction, and evaluation procedures. We report judge-model selection experiments and the corresponding hyperparameter settings, including reasoning effort and temperature, in Section~\ref{sec:judge_selection} and Table~\ref{tab:judge_selection}. Appendix~\ref{app:data_collection_prompts} provides prompt templates for intent and quality filtering, domain classification, task rewriting, and rubric-based evaluation in Chinese and English, together with the deep research agent system prompt. Figure~\ref{fig:dataset_profile} summarizes the task distribution and rubric statistics, while Table~\ref{tab:rubric_design_details} details the rubric dimensions, annotation fields, and writing requirements. Appendix~\ref{app:response_length} documents the evaluation window within a few days in August 2026 and the mean final-answer length for each model.

\bibliography{references}

@misc{openai2025deepresearch,
  title        = {Introducing deep research},
  author       = {OpenAI},
  year         = {2025},
  howpublished = {\url{https://openai.com/index/introducing-deep-research/}},
}

@misc{google2025deepresearch,
  author       = {{Gemini}},
  title        = {Gemini Deep Research},
  year         = {2025},
  howpublished = {\url{https://gemini.google/overview/deep-research/}},
}

@article{mialon2023gaia,
  title={GAIA: A Benchmark for General AI Assistants},
  author={Mialon, Gr{\'e}goire and Fourrier, Cl{\'e}mentine and Swift, Craig and Wolf, Thomas and LeCun, Yann and Scialom, Thomas},
  journal={arXiv preprint arXiv:2311.12983},
  year={2023}
}

@inproceedings{yao2024assistantbench,
  title={Assistantbench: Can web agents solve realistic and time-consuming tasks?},
  author={Yoran, Ori and Amouyal, Samuel Joseph and Malaviya, Chaitanya and Bogin, Ben and Press, Ofir and Berant, Jonathan},
  booktitle={Proceedings of the 2024 Conference on Empirical Methods in Natural Language Processing},
  pages={8938--8968},
  year={2024}
}

@article{wei2025browsecomp,
  title={BrowseComp: A Simple Yet Challenging Benchmark for Browsing Agents},
  author={Wei, Jason and others},
  journal={arXiv preprint arXiv:2504.12516},
  year={2025},
  url={https://arxiv.org/abs/2504.12516}
}

@inproceedings{zhou2024webarena,
 author = {Zhou, Shuyan and Xu, Frank F and Zhu, Hao and Zhou, Xuhui and Lo, Robert and Sridhar, Abishek and Cheng, Xianyi and Ou, Tianyue and Bisk, Yonatan and Fried, Daniel and Alon, Uri and Neubig, Graham},
 booktitle = {International Conference on Learning Representations},
 editor = {B. Kim and Y. Yue and S. Chaudhuri and K. Fragkiadaki and M. Khan and Y. Sun},
 pages = {15585--15606},
 title = {WebArena: A Realistic Web Environment for Building Autonomous Agents},
 url = {https://proceedings.iclr.cc/paper_files/paper/2024/file/4410c0711e9154a7a2d26f9b3816d1ef-Paper-Conference.pdf},
 volume = {2024},
 year = {2024}
}

@article{zhang2025deep,
  title={Deep research: A survey of autonomous research agents},
  author={Zhang, Wenlin and Li, Xiaopeng and Zhang, Yingyi and Jia, Pengyue and Wang, Yichao and Guo, Huifeng and Liu, Yong and Zhao, Xiangyu},
  journal={arXiv preprint arXiv:2508.12752},
  year={2025}
}

@inproceedings{du2026deepresearch,
  title={Deepresearch bench: A comprehensive benchmark for deep research agents},
  author={Du, Mingxuan and Xu, Benfeng and Zhu, Chiwei and Zhang, Licheng and Wang, Xiaorui and Mao, Zhendong},
  booktitle={International Conference on Learning Representations},
  volume={2026},
  pages={42414--42448},
  year={2026}
}

@inproceedings{sharma2026researchrubrics,
  title={Researchrubrics: A benchmark of prompts and rubrics for evaluating deep research agents},
  author={Sharma, Manasi and Zhang, Chen Bo Calvin and Bandi, Chaithanya and Wang, Clinton and Aich, Ankit and Nghiem, Huy and Rabbani, Tahseen and Htet, Ye and Jang, Brian and Basu, Sumana and others},
  booktitle={International Conference on Learning Representations},
  volume={2026},
  pages={90447--90472},
  year={2026}
}

@article{zhong2026draco,
  title={Draco: a cross-domain benchmark for deep research accuracy, completeness, and objectivity},
  author={Zhong, Joey and Zhang, Hao and Southern, Clare and Yang, Jeremy and Wang, Thomas and Jung, Kate and Zhang, Shu and Yarats, Denis and Ho, Johnny and Ma, Jerry},
  journal={arXiv preprint arXiv:2602.11685},
  year={2026},
  url={https://arxiv.org/abs/2602.11685}
}

@article{singh2025code,
  title={Code researcher: Deep research agent for large systems code and commit history},
  author={Singh, Ramneet and Joel, Sathvik and Mehrotra, Abhav and Wadhwa, Nalin and Bairi, Ramakrishna B and Kanade, Aditya and Natarajan, Nagarajan},
  journal={arXiv preprint arXiv:2506.11060},
  year={2025}
}

@article{chen2025medbrowsecomp,
  title={Medbrowsecomp: Benchmarking medical deep research and computer use},
  author={Chen, Shan and Moreira, Pedro and Xiao, Yuxin and Schmidgall, Sam and Warner, Jeremy and Aerts, Hugo and Hartvigsen, Thomas and Gallifant, Jack and Bitterman, Danielle S},
  journal={arXiv preprint arXiv:2505.14963},
  year={2025}
}

@article{zhu2025findeepresearch,
  title={Findeepresearch: Evaluating deep research agents in rigorous financial analysis},
  author={Zhu, Fengbin and Ng, Xiang Yao and Liu, Ziyang and Liu, Chang and Zeng, Xianwei and Wang, Chao and Tan, Tianhui and Yao, Xuan and Shao, Pengyang and Xu, Min and others},
  journal={arXiv preprint arXiv:2510.13936},
  year={2025}
}

@inproceedings{wong2026widesearch,
 author = {Wong, Ryan and Wang, Jiawei and zhao, Junjie and Chen, Li and Gao, Yan and Zhanglong and Zhou, Xuan and Wang, Zuo and Xiang, Kai and Zhang, Ge and Huang, Wenhao and WANG, YANG and Wang, Ke},
 booktitle = {International Conference on Learning Representations},
 editor = {C. Vondrick and B. Hariharan and C. Raffel and L. Pinto and D. Yang and A. Faust},
 pages = {10012--10086},
 title = {WideSearch: Benchmarking Agentic Broad Info-Seeking},
 url = {https://proceedings.iclr.cc/paper_files/paper/2026/file/1119587863e78451f080da2a768c4935-Paper-Conference.pdf},
 volume = {2026},
 year = {2026}
}

@article{li2025reportbench,
  title={Reportbench: Evaluating deep research agents via academic survey tasks},
  author={Li, Minghao and Zeng, Ying and Cheng, Zhihao and Ma, Cong and Jia, Kai},
  journal={arXiv preprint arXiv:2508.15804},
  year={2025}
}

@article{gou2025mind2web2,
  title={Mind2Web 2: Evaluating Agentic Search with Agent-as-a-Judge},
  author={Gou, Boyu and Huang, Zanming and Ning, Yuting and Gu, Yu and Lin, Michael and Qi, Weijian and Kopanev, Andrei and Yu, Botao and Guti{\'e}rrez, Bernal Jim{\'e}nez and Shu, Yiheng and Song, Chan Hee and Wu, Jiaman and Chen, Shijie and Moussa, Hanane Nour and Zhang, Tianshu and Xie, Jian and Li, Yifei and Xue, Tianci and Liao, Zeyi and Zhang, Kai and Zheng, Boyuan and Cai, Zhaowei and Rozgic, Viktor and Ziyadi, Morteza and Sun, Huan and Su, Yu},
  journal={arXiv preprint arXiv:2506.21506},
  year={2025},
  url={https://arxiv.org/abs/2506.21506}
}

@inproceedings{wang2025liveresearchbench,
  title={Liveresearchbench: A live benchmark for user-centric deep research in the wild},
  author={Wang, Jiayu and Ming, Yifei and Dulepet, Riya and Chen, Qinglin and Xu, Austin and Ke, Zixuan and Sala, Frederic and Albarghouthi, Aws and Xiong, Caiming and Joty, Shafiq},
  booktitle={International Conference on Learning Representations},
  volume={2026},
  pages={95096--95143},
  year={2026}
}

@article{wang2026deepresearcheval,
  title={{DeepResearchEval}: An Automated Framework for Deep Research Task Construction and Agentic Evaluation},
  author={Wang, Yibo and Wang, Lei and Deng, Yue and Wu, Keming and Xiao, Yao and Yao, Huanjin and Kang, Liwei and Ye, Hai and Jing, Yongcheng and Bing, Lidong},
  journal={arXiv preprint arXiv:2601.09688},
  year={2026},
  url={https://arxiv.org/abs/2601.09688}
}

@article{jin2025search,
  title={Search-r1: Training llms to reason and leverage search engines with reinforcement learning},
  author={Jin, Bowen and Zeng, Hansi and Yue, Zhenrui and Yoon, Jinsung and Arik, Sercan and Wang, Dong and Zamani, Hamed and Han, Jiawei},
  journal={arXiv preprint arXiv:2503.09516},
  year={2025}
}

@misc{kimiResearcher,
  title={{Kimi-Researcher}: End-to-End {RL} Training for Emerging Agentic Capabilities},
  author={{Moonshot AI}},
  year={2025},
  howpublished={Technical report},
  url={https://moonshotai.github.io/Kimi-Researcher/}
}

@article{team2025tongyi,
  title={Tongyi deepresearch technical report},
  author={Team, Tongyi DeepResearch and Li, Baixuan and Zhang, Bo and Zhang, Dingchu and Huang, Fei and Li, Guangyu and Chen, Guoxin and Yin, Huifeng and Wu, Jialong and Zhou, Jingren and others},
  journal={arXiv preprint arXiv:2510.24701},
  year={2025}
}

@misc{googledeepmind2026gemini31pro,
  author       = {{Google DeepMind}},
  title        = {{Gemini 3.1 Pro}},
  year         = {2026},
  month        = feb,
  howpublished = {Model card},
  url          = {https://deepmind.google/models/model-cards/gemini-3-1-pro/},
  note         = {Published February 19, 2026. Accessed September 20, 2026}
}

@article{peng2026can,
  title={Can {LLM-as-a-Judge} Reliably Verify Rubrics in Agentic Scenarios?},
  author={Peng, Yangda and Qi, Yunjia and Peng, Hao and Xia, Haotian and He, Guanzhong and Shi, Xintong and Xuan, Richeng and Lu, Songyuanyi and Liu, Yixian and Hu, Zhichao and others},
  journal={arXiv preprint arXiv:2606.29920},
  year={2026}
}

@inproceedings{DBLP:conf/iclr/YaoZYDSN023,
  author       = {Shunyu Yao and
                  Jeffrey Zhao and
                  Dian Yu and
                  Nan Du and
                  Izhak Shafran and
                  Karthik R. Narasimhan and
                  Yuan Cao},
  title        = {ReAct: Synergizing Reasoning and Acting in Language Models},
  booktitle    = {The Eleventh International Conference on Learning Representations,
                  {ICLR} 2023, Kigali, Rwanda, May 1-5, 2023},
  publisher    = {OpenReview.net},
  year         = {2023},
  url          = {https://openreview.net/forum?id=WE\_vluYUL-X},
  bibsource    = {dblp computer science bibliography, https://dblp.org}
}
\bibliographystyle{iclr2027_conference}

\appendix


\newpage
\section{Rubric Design Details}
\label{app:rubric_design}

Table~\ref{tab:rubric_design_details} expands the overview in Section~\ref{sec:rubric_design} with the annotation fields, authoring requirements, and illustrative examples.

\begin{table}[h]
\caption{Rubric assessment dimensions, record fields, and authoring requirements. Examples are illustrative and refer to fictional subscription plans A and B.}
\label{tab:rubric_design_details}
\centering
\fontsize{8}{9.5}\selectfont
\setlength{\tabcolsep}{3pt}
\renewcommand{\arraystretch}{1.05}
\begin{tabular}{@{}>{\raggedright\arraybackslash}p{0.22\linewidth}>{\raggedright\arraybackslash}p{0.33\linewidth}>{\raggedright\arraybackslash}p{\dimexpr0.45\linewidth-12pt\relax}@{}}
\toprule
\textbf{Aspect} & \textbf{Definition or requirement} & \textbf{Illustrative rubric or value} \\
\midrule
\multicolumn{3}{@{}l}{\textbf{A. Assessment dimensions}} \\
\midrule
Instruction following & Fulfillment of explicit user requests. & The response should recommend a choice between Plan A and Plan B. \\
Information coverage & Coverage of the requested entities, attributes, and scope. & The response should report the monthly subscription fees of both Plan A and Plan B. \\
Factual accuracy & Correctness of concrete facts and numerical targets. & The response should state that Plan A costs USD 20 per month as of 30 June 2026. \\
Evidence and citation quality & Required source identity, attribution, and temporal context. & The response should identify Plan A's official pricing page as the source of its monthly fee. \\
Reasoning and analysis & An explicit connection between evidence and conclusions. & The response should explain how Plan A's lower fee supports the recommendation under a USD 25 monthly budget. \\
Task success & A conclusion that resolves the user's decision or question. & The response should conclude that Plan A meets the USD 25 monthly budget. \\
Presentation & Compliance with explicitly requested structure and format. & The response should present Plan A and Plan B in a comparison table. \\
\midrule
\multicolumn{3}{@{}l}{\textbf{B. Annotation fields}} \\
\midrule
\texttt{criterion} & A complete, independently judgeable acceptance statement. & ``The response should \ldots'' followed by the concrete requirement. \\
\texttt{type} & Whether the judgment is objectively checkable or subjective. & \textsc{Objective} / \textsc{Subjective}. Verifiable prices are objective; stylistic preferences are subjective. \\
\texttt{necessity} & Whether the requirement is stated in the query or reasonably implied. & \textsc{Explicit} / \textsc{Implicit}. A requested price comparison is explicit. \\
\texttt{priority} & Qualitative importance label retained as input to the judge. & \textsc{Must Have} / \textsc{Nice to Have}; the score uses the separately recorded numerical weight. \\
\texttt{weight} & Relative importance of the requirement within its task, assigned by an annotator. & A floating-point value from $0.0$ to $1.0$ with one decimal place; larger values indicate greater importance. \\
\midrule
\multicolumn{3}{@{}l}{\textbf{C. Rubric-writing requirements}} \\
\midrule
Atomic and non-redundant & One independently testable requirement per item; no duplicate checks. & Separate a price check from a recommendation check; merge repeated checks of the same price. \\
Self-contained & Identify the entities and scope needed for standalone judgment. & Name ``Plan A as of 30 June 2026'' instead of ``the former'' or ``currently.'' \\
Verified factual targets & Encode the checked fact needed for grading without fresh retrieval. & Supply the verified price in the rubric; do not ask the judge to discover it online. \\
Scope and tolerance & Specify units, dates, ranges, and permitted approximations where relevant. & For a request of about 100 words, specify an accepted range of 80--120 words. \\
Affirmative, operational wording & State the desired behavior positively and avoid vague quality adjectives. & Write ``The response should distinguish monthly from annual billing,'' not ``be accurate and detailed.'' \\
Valid answer alternatives & Preserve reasonable solution diversity for open-ended tasks. & Make example-specific recommendations optional unless explicitly required; assess the comparison independently. \\
\bottomrule
\end{tabular}
\end{table}

\clearpage
\raggedbottom
\section{Data Collection and Evaluation Prompt Templates}
\label{app:data_collection_prompts}
We provide English translations of the filtering, domain classification, and rewriting prompts associated with Section~\ref{sec:data_collection}. The filtering template presents three selected judgments from the original combined prompt, renumbered as 1, 2, and 3. Output keys remain unchanged, while quality labels and category names are translated into English. Placeholders indicate the input query.
We also provide the original Chinese and English judge prompt templates used for rubric evaluation in Section~\ref{sec:evaluation_methodology}, including the \texttt{priority} field. The judge determines rubric satisfaction; the annotator-assigned weights are applied separately during score aggregation.
The search-agent system prompt is included in Appendix~\ref{app:agent_system_prompt}.

\Needspace{10\baselineskip}
\subsection{Intent and Quality Filtering}
\label{app:filter_prompt}
The following template combines task quality, search relevance, and implicit event-dependence assessment. Its time-dependence flag distinguishes underspecified event premises from general requests for up-to-date information.
\begin{prompttemplate}{Intent and Quality Filtering}
\begin{PromptTemplateVerbatim}
You are a search/research task reviewer. Make the following three judgments
about the task in a single response. Return strict JSON.

[Judgment 1: quality_level (three task-quality levels)]
- "high": Clear, complete, and actionable, with an explicit goal and scope;
  the task is adequately specified and its information needs are concrete.
- "medium": Generally actionable, with minor ambiguity or local omissions,
  but the overall intent is understandable.
- "low": Ambiguous, truncated, missing essential information, merely a
  follow-up fragment, full of disruptive spelling or grammatical errors,
  meaningless, incoherent after rewriting, clearly abandoned, or only a
  debugging query.

[Judgment 2: is_unrelated_to_search]
Related to search/research:
- Information retrieval, news aggregation, market analysis, industry research.
- Data collection, competitor research, policy or regulatory lookup.
- Academic literature searches; background research on companies, people,
  or events.
- Queries for public prices, market quotations, stocks, or exchange rates.
Unrelated to search/research:
- Pure code writing, refactoring, or debugging.
- Pure file reading, writing, or editing without external information search.
- Pure mathematical calculation or logical reasoning.
- Pure creative writing without external information needs.
- Casual conversation, greetings, or role-play.
- Pure translation without prior search.

[Judgment 3: is_time_dependent]
Focus on whether the task presupposes a specific recent event or change
in state.
Typical true cases (to filter out):
- "Why has X become more expensive recently?" / "Why has X fallen recently?"
  These presuppose a particular recent price change.
- "Has X received an update or promotion recently? Its ranking has risen
  so much." This presupposes a particular recent ranking change.
- "Why has Company A's stock price surged recently?" This presupposes a
  particular recent price event.
- "What major moves has Product X made recently?" This presupposes a
  particular development already perceived by the user.
- A particular, unstated recent change is an implicit premise.
- Without having recently observed the event, the task is unclear or its
  answer may not match the user's expectation.
- The answer depends heavily on an event within a particular time window.
Typical false cases:
- "Search today's technology news" / "Summarize AI news over the past
  24 hours": general time-sensitive searches without a presupposed event.
- "What are the latest GPT models?" / "Find the latest data-center policies":
  general queries without a presupposed event.
- "Find Company X's stock price today" / "Today's exchange rates": real-time
  information queries without a presupposed event.
- "Industry X's development trends over the past three years": long-term
  trend analysis without dependence on a particular event.
Key question: Does the task depend on a specific recent event that the user
assumes but does not explicitly describe? If so, return true. For general
time-sensitive searches, return false.

[Output]
Return only a valid JSON object with exactly the following fields.
Use "high", "medium", or "low" for quality_level and JSON booleans for the
two flags. Give a one-sentence reason for each judgment; "ok" is acceptable
when a flag is false. The object below illustrates the required format:
{
  "quality_level": "high",
  "quality_reason": "One-sentence explanation.",
  "is_unrelated_to_search": false,
  "unrelated_reason": "ok",
  "is_time_dependent": false,
  "time_dependent_reason": "ok"
}

[Task]
{query}
\end{PromptTemplateVerbatim}
\end{prompttemplate}

\Needspace{10\baselineskip}
\subsection{Domain Classification}
\label{app:domain_prompt}
This template is the English-query version of the seven-category classifier. It assigns one category according to the query's core purpose.
\begin{prompttemplate}{Domain Classification}
\begin{PromptTemplateVerbatim}
You are a labeling lead for a search benchmark. Classify the given English
user query into exactly ONE of the seven categories below. Output the
category's English name.

# Seven categories
1. Financial market analysis: stock, futures, fund, foreign-exchange, or
   index valuation; trends; stock picking; arbitrage; and market moves.
2. Academic and professional literature: finding papers, patents, or reviews;
   investigating advanced technical principles; and peer-reviewed sources.
3. Company and competitor research: profiling specific companies or products;
   competitor comparisons; industry landscapes; buyers, importers, or
   distributors; tenders; and industry reports.
4. Industry news and trend aggregation: recent news, trends, or prominent
   events in a field; time-sensitive aggregation across sources; and research
   on market outlooks, prospects, or business models.
5. Policy, regulation and government data: policy provisions, regulations,
   government statistics, administrative penalties, budgets, and employment
   outcome data.
6. Technical tools and developer resources: finding tools, software, source
   code, or APIs; how-to guides; and developer-resource lookup.
7. Everyday facts: everyday factual information, general knowledge,
   single-point lookups, history, biographies, and sports records.

# Rules
- Classify by the CORE PURPOSE, not the surface topic.
- Finding buyers, importers, or company information -> Company and
  competitor research; overall field news/trends -> Industry news and
  trend aggregation.
- Finding tools, APIs, or code -> Technical tools and developer resources;
  finding papers, patents, or academic principles -> Academic and
  professional literature.

# Output
Return only a valid JSON object, without Markdown code fences or additional
explanation. Use one of the seven category names and a reason of at most
10 words:
{
  "taxonomy": "<one of the seven English category names>",
  "reason": "<short reason, at most 10 words>"
}

# Current query
{query}
\end{PromptTemplateVerbatim}
\end{prompttemplate}

\Needspace{10\baselineskip}
\subsection{Task Rewriting}
\label{app:rewrite_prompt}
This translation of \texttt{REWRITE\_PROMPT} removes operational requirements while preserving the information-seeking request and the original query language. Product names in this template are replaced with generic descriptions.
\begin{prompttemplate}{Task Rewriting}
\begin{PromptTemplateVerbatim}
You are a data-cleaning assistant. Transform an original user query into
a version that contains only search/research tasks.

[Input]
An original user query's main purpose is
search/research, but it may contain engineering operations and local-file
dependencies unrelated to that purpose.

[Goal]
Remove non-search content with minimal intervention. Produce a search/research
task that requires neither local files nor engineering operations.

[Remove]
Remove matching paragraphs, steps, or sections in full, subject to the
mixed-sentence rule below.
1. Saving files: generating or saving .md/.docx/.json/.csv/.html files,
   file-naming rules, and storage paths.
2. Explicit skill/tool calls: using a named skill, update_memory, image_gen,
   deliver_attachments, preview_url, the open command, web_search, web_fetch,
   or other explicitly named built-in tools.
3. Knowledge-base archiving: archiving in a knowledge-management service,
   writing to a knowledge base, or reading, writing, and updating
   memory.md / MEMORY.md.
4. Version control and delivery: git add/commit/push, messaging-service
   webhooks, email delivery, and PowerShell-based delivery.
5. Service/script calls: curl, scp, osascript, PowerShell, bash, Python,
   text-to-speech, podcast generation, and related commands or code blocks.
6. Automation-framework context: "read memory -> execute -> update memory"
   workflows, idempotency checks, and scheduling directives such as
   "run on weekends too" or "do not skip."
7. Engineering-specific output definitions: JSON schema fields, field order,
   file-name formats, storage paths, and other implementation details.
8. Local-file dependencies: attached-file references such as
   @/Users/X/.../doc.pdf; requests to read local, state, template, or
   configuration files or historical logs; and any step that requires a
   local file as input.
9. Output file/markup declarations, but not content-organization requirements:
   "output in Markdown"; requests to output .md/.html/.docx/.json/.csv files;
   and instructions about Markdown heading or table syntax.

[Preserve verbatim]
1. Search topics: industries, companies, news, papers, policies, market data,
   and other information targets.
2. Recency expressions such as "today", "the past 24 hours", "latest", or
   "the previous day."
3. Information sources, keyword lists, coverage, and priorities.
4. The original author's voice, sentence structure, wording, and tone.
5. Output-length and item-count constraints, such as "1,000-1,500 Chinese
   characters" or "3-5 items."
6. Content-organization requirements such as a structured daily report,
   importance levels, sections, topic blocks, or Top N lists. Distinguish
   these from the file/markup declarations in removal rule 9.

[Processing principles]
- This is removal, not free rewriting: retain original wording wherever
  possible.
- Exception for minimal edits: if a remaining sentence directly depends
  on a removed file, make the smallest edit necessary for coherence.
  For example, when "compare the latest data with the previous record"
  refers to a removed state file, remove references such as "this document",
  "the previous record", or "in the template", or simplify the request
  to "find the latest data." Keep each edit within a single sentence.
- If a sentence mixes search content with engineering operations or file
  dependencies, remove only the engineering/file-related part and retain
  the search content.
- Remove headings left without content, such as "## Output requirements."
- If nothing remains because the entire query consists of engineering
  operations, return an empty string.
- Keep the language of the original query (Chinese or English).

[Output]
Return the rewritten query directly, without explanations or Markdown
code fences.

[Original query]
{user_query}
\end{PromptTemplateVerbatim}
\end{prompttemplate}

\Needspace{10\baselineskip}
\subsection{English Judge Prompt}
\label{app:judge_prompt_en}
For English tasks, the judge uses the following prompt with the same inputs and output fields.
\begin{prompttemplate}{English Judge Prompt}
\begin{PromptTemplateVerbatim}
You are a strict grader. Below you are given a research task's [Question], the [Model Response], and one [Criterion]. Judge whether the model response satisfies this criterion.

[Question]
{question}

[Model Response]
{answer}

[Criterion]
- criterion: {criterion}
- type: {type} (Objective = objectively verifiable fact / Subjective = subjective judgment)
- necessity: {necessity} (Explicit = explicitly required by the question / Implicit = reasonably implied expectation)
- priority: {priority} (Must Have = must be satisfied / Nice to Have = bonus)

Output ONLY a strict JSON object, with nothing else (do NOT wrap it in a markdown ``` code block):
{"satisfied": true or false, "reason": "reason in English, <= 30 words"}

Judging principles:
- satisfied=true if and only if: the model response contains content that clearly covers what this criterion requires.
- satisfied=false if: the response does not address this criterion, or the information given does not match / is wrong.
- If the criterion says "should state X / should list Y / should give Z", the model must actually state / list / give the corresponding content to pass.
- Do not be lenient just because the response is long or well-structured; and do not automatically fail it just because it is short.
\end{PromptTemplateVerbatim}
\end{prompttemplate}

\Needspace{10\baselineskip}
\subsection{Chinese Judge Prompt}
\label{app:judge_prompt_zh}
For Chinese tasks, the judge receives the following prompt for each rubric. The placeholders supply the question, model response, rubric text, and annotation fields. The output is a binary satisfaction label with a brief reason. Both language versions retain the input placeholders and leave numerical weighting to the aggregation step.
\begin{prompttemplate}{Chinese Judge Prompt}
\begin{judgechinese}
\begin{PromptTemplateVerbatim}[formatcom=\color{black}\judgechinesefont]
你是一个严格的评分员。下面给你一个研究任务的【题目】、【模型答复】和一条【评分标准】。请判断该模型答复是否满足这条评分标准。

【题目】
{question}

【模型答复】
{answer}

【评分标准】
- criterion: {criterion}
- type: {type}（Objective=客观事实可核实 / Subjective=主观判断）
- necessity: {necessity}（Explicit=题目明示要求 / Implicit=隐含期望）
- priority: {priority}（Must Have=必须满足 / Nice to Have=锦上添花）

请只输出一个严格的 JSON 对象，无任何其他内容（不要包 markdown ``` 代码块）：
{"satisfied": true 或 false, "reason": "≤80字中文理由"}

判断原则：
- satisfied=true 当且仅当：模型答复中**有内容明确覆盖该评分标准的要求**。
- satisfied=false 若：答复未涉及该标准、或所给信息与标准不符 / 错误。
- 标准里若说"应说明 X / 应列出 Y / 应给出 Z"，模型必须真的说出/列出/给出对应内容才算通过。
- 不要因为答复整体写得长 / 结构好就放宽；也不要因为篇幅短就一律打 false。
\end{PromptTemplateVerbatim}
\end{judgechinese}
\end{prompttemplate}

\Needspace{10\baselineskip}
\subsection{Deep Research Agent System Prompt}
\label{app:agent_system_prompt}
The following system prompt guides the search agents. The placeholder \texttt{\{today\_date\}} supplies the current date at execution time.
\begin{prompttemplate}{Deep Research Agent System Prompt}
\begin{PromptTemplateVerbatim}
You are a highly capable search agent that handles research tasks of any depth or breadth—from deep multi-step investigations to broad cross-entity surveys. End with a clear, calibrated answer grounded in the evidence you actually used.

The current date is {today_date}.

# Doing research tasks
- Treat the user's stated dates, time windows, and premises as fixed task parameters, and execute the task exactly as scoped even if a date seems past or the premise seems off. The stated date is a hard cutoff: every fact, figure, price, and event in your answer must be dated at or before it—do not use newer information you retrieved while searching. You must complete the task under the given constraints rather than asking the user to reconfirm before answering.
- Think carefully through the task, and you MUST use tools to help your reasoning before solving it. Never ask for human help.
- Before finalizing, re-read the user's question and list every explicit sub-requirement it contains; make sure your answer addresses each one specifically. A research task is scored on whether you surface the exact facts asked for (names, dates, figures, specific attributes, edge cases).
\end{PromptTemplateVerbatim}
\end{prompttemplate}

\section{Response Length and Evaluation Timing}
\label{app:response_length}
Table~\ref{tab:response_length} reports the mean final-answer length per task in characters for the 12 models in Table~\ref{tab:main_results}, averaged over the available runs. All models listed in the table are evaluated within a few days of one another in \textbf{August 2026}.

\begin{table}[htbp]
\caption{Mean final-answer length per task. Length is measured in characters, not tokens.}
\label{tab:response_length}
\centering
\small
\setlength{\tabcolsep}{10pt}
\renewcommand{\arraystretch}{1.08}
\begin{tabular}{@{}lr@{}}
\toprule
\textbf{Model} & \textbf{Mean answer length (characters)} \\
\midrule
GPT-5.6 Sol & 23,679 \\
Claude Opus 5 & 12,264 \\
Hy4 Preview & 9,959 \\
GLM-5.3 & 10,381 \\
Qwen3.8-Max & 9,166 \\
DeepSeek-V4 Pro & 9,189 \\
Kimi-K3 & 7,832 \\
Doubao-Seed-2.1-Pro & 12,089 \\
Claude Opus 4.8 & 7,576 \\
Hy3 & 6,522 \\
GLM-5.2 & 8,613 \\
Gemini-3.6-Flash & 10,256 \\
\bottomrule
\end{tabular}
\end{table}

\clearpage
\section{Model Responses and Rubric Judgments}
\label{app:representative_cases}
We present four examples from recorded model trajectories. Each case begins with a brief task summary, followed by selected response excerpts and two recorded rubric judgments---one Pass and one Fail. Table rows are reformatted as labeled fields, and editorial explanations are identified separately from model text. Scores and passed-rubric counts refer to the complete report and full rubric set. The original task prompts are omitted; quoted model claims are not independently verified here.

\Needspace{14\baselineskip}
\subsection{Case 1: Dify Workflow and Chatflow}
\label{app:case_dify}
\noindent\textbf{Model:} Hy4 Preview \quad \textbf{Language:} Chinese\\
\textbf{Domain:} Technical tools and developer resources\\
\textbf{Task type:} Tool comparison and workflow selection\\
\textbf{Full-report evaluation:} 14/18 rubrics passed; score 60.38\%.

\noindent\textbf{Task summary.} A team moving knowledge-base Q\&A and back-office automation to Dify needs to choose between Workflow and Chatflow. The report should compare their execution modes, start nodes, and use cases using official documentation and dated release notes.\par\smallskip
\begin{CJK*}{UTF8}{gbsn}
\begin{prompttemplate}{Response excerpts}
\normalfont\fontsize{9}{11.5}\selectfont\raggedright
\noindent\textbf{1. Workflow execution (excerpt).} Workflow \& Chatflow 页："A Workflow runs once from start to finish. It takes an input, processes it through the flow, and returns a result."\par\smallskip
\noindent\textbf{2. Start-node options (reformatted table row).} \par\smallskip
\noindent\textbf{Workflow:} User Input 或 Trigger\par\smallskip
\noindent\textbf{Chatflow:} User Input（仅此一种）\par\smallskip
\noindent\textbf{Source label in the model's table:} Workflow \& Chatflow 对比表\par\smallskip
\end{prompttemplate}
\noindent\textbf{Case explanation (editorial).} The response describes single-run execution and lists both start-node options. It does not state that User Input and Trigger are mutually exclusive.\par\smallskip
\begin{prompttemplate}{Two recorded rubric judgments}
\normalfont\fontsize{9}{11.5}\selectfont\raggedright
\noindent\textbf{R4 -- Pass}\par\smallskip
\noindent\textbf{Rubric:} 回复应定义 Workflow 为单轮执行模式：接收输入→流经节点处理→返回结果，运行一次即结束\par\smallskip
\noindent\textbf{Recorded reason:} 答复在定义与定位部分明确引用官方原话，说明Workflow处理单轮任务，接收输入、通过流程处理并返回结果，运行一次即结束，完全符合标准。\par\smallskip
\par\medskip
\noindent\textbf{R6 -- Fail}\par\smallskip
\noindent\textbf{Rubric:} 回复应说明 Workflow 的开始节点可以是 User Input 或 Trigger（定时/Webhook/插件事件），且二者互斥，不可同时使用\par\smallskip
\noindent\textbf{Recorded reason:} 模型答复说明了Workflow的开始节点可以是User Input或Trigger，但未提及二者互斥、不可同时使用。\par\smallskip
\end{prompttemplate}
\end{CJK*}

\clearpage
\subsection{Case 2: Winter-Wheat Literature Screening}
\label{app:case_winter_wheat}
\noindent\textbf{Model:} Hy4 Preview \quad \textbf{Language:} Chinese\\
\textbf{Domain:} Academic and professional literature\\
\textbf{Task type:} Literature retrieval and eligibility screening\\
\textbf{Full-report evaluation:} 11/15 rubrics passed; score 57.53\%.

\noindent\textbf{Task summary.} A research group needs English-language papers in Q1 (top-quartile) journals on winter-wheat ridge/furrow or raised-bed cultivation. It requests search terms, verified bibliographic details, and journal-quartile checks against Web of Science and Journal Citation Reports (JCR).\par\smallskip
\begin{CJK*}{UTF8}{gbsn}
\begin{prompttemplate}{Response excerpts}
\normalfont\fontsize{9}{11.5}\selectfont\raggedright
\noindent\textbf{1. Search strategy (excerpt).} \par\smallskip
\noindent\textbf{Reading guide (editorial).} The query combines crop terms with cultivation terms. Each TS=(...) group supplies search terms; AND requires both groups, while OR permits alternatives within a group.\par\smallskip
\begin{Verbatim}[fontsize=\fontsize{8}{10}\selectfont,breaklines=true,breakautoindent=false,breakindent=0pt]
TS=("winter wheat" OR "Triticum aestivum")
AND TS=("ridge-furrow*" OR "ridge and furrow"
        OR "furrow-ridge" OR "ridge tillage"
        OR "ridge planting")
\end{Verbatim}
\noindent\textbf{2. A paper included in the model's list (reformatted row).} \par\smallskip
\noindent\textbf{Title:} Distribution of soil carbon and grain yield of \textbf{spring wheat} under a permanent raised bed planting system in an arid area of northwest China\par\smallskip
\noindent\textbf{Year / author:} 2016 / Ma Z. 等5人\par\smallskip
\noindent\textbf{Journal:} Soil Tillage Res. 163, 274--281\par\smallskip
\noindent\textbf{DOI:} \nolinkurl{10.1016/j.still.2016.05.010}\par\smallskip
\end{prompttemplate}
\noindent\textbf{Case explanation (editorial).} The requested crop is winter wheat, but the listed paper's title explicitly says spring wheat. A relevant search query does not ensure that every selected paper meets the crop requirement.\par\smallskip
\begin{prompttemplate}{Two recorded rubric judgments}
\normalfont\fontsize{9}{11.5}\selectfont\raggedright
\noindent\textbf{R1 -- Pass}\par\smallskip
\noindent\textbf{Rubric:} 回复的检索词组中应包含"winter wheat"，并配合"ridge-furrow"、"furrow-ridge"、"raised bed"、“ridge tillage”进行搭配\par\smallskip
\noindent\textbf{Recorded reason:} 模型在检索式中明确给出了包含'winter wheat'（或wheat）与'ridge-furrow'、'furrow-ridge'、'raised bed'、'ridge tillage'等词的布尔逻辑搭配方法。\par\smallskip
\par\medskip
\noindent\textbf{R11 -- Fail}\par\smallskip
\noindent\textbf{Rubric:} 回复应确保所列论文的研究对象必须包含冬小麦（winter wheat），不应包含春小麦或其他作物\par\smallskip
\noindent\textbf{Recorded reason:} 模型列出的文献清单中包含了一篇明确研究春小麦（spring wheat）的论文，违反了不应包含春小麦的评分标准。\par\smallskip
\end{prompttemplate}
\end{CJK*}

\clearpage
\subsection{Case 3: A Time-Bounded Repository Profile}
\label{app:case_repository}
\noindent\textbf{Model:} DeepSeek-V4 Pro \quad \textbf{Language:} English\\
\textbf{Domain:} Technical tools and developer resources\\
\textbf{Task type:} Time-bounded repository profiling and fact verification\\
\textbf{Full-report evaluation:} 14/26 rubrics passed; score 41.55\%.

\noindent\textbf{Task summary.} A course-resource tracker needs a profile of rasbt/LLMs-from-scratch as of July 20, 2026, covering popularity, maintenance, license/languages, README resources, and unsupported scale claims. The time cutoff applies to all reported information.\par\smallskip
\begin{prompttemplate}{Response excerpts}
\normalfont\fontsize{9}{11.5}\selectfont\raggedright
\noindent\textbf{1. Date disclosure (excerpt).} Dating caveat up front. GitHub's REST API only returns a live snapshot; my API reads came back stamped 2026-08-21 (about a month after your 2026-07-20 cutoff).\par\smallskip
\noindent\textbf{2. Reported metrics (reformatted table rows).} \par\smallskip
\noindent\textbf{Watchers (GitHub field):} 103,146\par\smallskip
\noindent\textbf{Model's qualifier:} API field; equals stars on modern GitHub --- do not double-count\par\smallskip
\noindent\textbf{Subscribers ("watching" for notifications):} 838\par\smallskip
\noindent\textbf{Model's qualifier:} API subscribers\_count, live 2026-08-21\par\smallskip
\end{prompttemplate}
\noindent\textbf{Case explanation (editorial).} The response distinguishes watchers from subscribers, satisfying the metric-definition criterion. However, its own disclosure dates the API values to August 21, after the requested July 20 cutoff.\par\smallskip
\begin{prompttemplate}{Two recorded rubric judgments}
\normalfont\fontsize{9}{11.5}\selectfont\raggedright
\noindent\textbf{R5 -- Pass}\par\smallskip
\noindent\textbf{Rubric:} The response should distinguish between watchers\_count from the GitHub API (e.g., equal to stars at 99,065) and actual subscribers/watchers (e.g., approximately 813), and label the meaning of each metric separately.\par\smallskip
\noindent\textbf{Recorded reason:} The response explicitly distinguishes between the GitHub API 'watchers' field (which equals stars) and actual subscribers, labeling the meaning of each metric clearly.\par\smallskip
\par\medskip
\noindent\textbf{R24 -- Fail}\par\smallskip
\noindent\textbf{Rubric:} The response should make sure all information is up to July 20, 2026.\par\smallskip
\noindent\textbf{Recorded reason:} The model explicitly includes information and data from after July 20, 2026, such as commits from August and API data from August 21, 2026.\par\smallskip
\end{prompttemplate}

\clearpage
\subsection{Case 4: UWB Literature and Method Comparison}
\label{app:case_uwb}
\noindent\textbf{Model:} DeepSeek-V4 Pro \quad \textbf{Language:} English\\
\textbf{Domain:} Academic and professional literature\\
\textbf{Task type:} Reference verification and methodological synthesis\\
\textbf{Full-report evaluation:} 14/18 rubrics passed; score 62.07\%.

\noindent\textbf{Task summary.} A researcher writing a UWB indoor-localization paper needs verified literature and dataset sources as of July 2026. The review should cover classical localization algorithms, neural approaches, and the UTIL dataset, with reliable references and explanations of the methods.\par\smallskip
\begin{prompttemplate}{Response excerpts}
\normalfont\fontsize{9}{11.5}\selectfont\raggedright
\noindent\textbf{Reference category:} Chan / WLS foundations (TDOA)\par\smallskip
\noindent\textbf{Cited paper (excerpt):} Y. T. Chan and K. C. Ho, "A simple and efficient estimator for hyperbolic location," IEEE Transactions on Signal Processing, vol. 42, no. 8, pp. 1905--1915, Aug. 1994. DOI: 10.1109/78.301830.\par\smallskip
\noindent\textbf{Model's explanation (excerpt):} This is the canonical closed-form, two-stage weighted least squares (WLS) TDOA solver; cite it for your Chan/WLS baseline.\par\smallskip
\end{prompttemplate}
\noindent\textbf{Case explanation (editorial).} The response supplies the requested Chan/WLS reference. The recorded judge nevertheless finds that the full report omits the required comparison of Chan-WLS, Taylor iteration, and constrained least squares.\par\smallskip
\begin{prompttemplate}{Two recorded rubric judgments}
\normalfont\fontsize{9}{11.5}\selectfont\raggedright
\noindent\textbf{R4 -- Pass}\par\smallskip
\noindent\textbf{Rubric:} The response should present the core foundational literature for classic Chan/WLS localization: Y. T. Chan, K. C. Ho. 1994. A Simple and Efficient Estimator for Hyperbolic Location.\par\smallskip
\noindent\textbf{Recorded reason:} The response explicitly includes the required 1994 paper by Y. T. Chan and K. C. Ho on the estimator for hyperbolic location under the Chan/WLS foundations section.\par\smallskip
\par\medskip
\noindent\textbf{R6 -- Fail} (implicit requirement)\par\smallskip
\noindent\textbf{Rubric:} The response should distinguish the differentiated positioning principles of the Chan-WLS algorithm, Taylor iterative algorithm, and constrained least squares algorithm when citing references.\par\smallskip
\noindent\textbf{Recorded reason:} The response does not mention the constrained least squares algorithm or distinguish the positioning principles of the three algorithms as required.\par\smallskip
\end{prompttemplate}

\clearpage
\section{Bilingual Task and Rubric Examples}
\label{app:task_rubric_examples}
We present one Chinese and one English task from \benchmark, with selected rubrics, respectively. Task and rubric wording is reproduced from the dataset, and rubric numbers follow their original order. We display \texttt{type}, \texttt{necessity}, and \texttt{priority}; numerical weights are omitted.

\subsection{Chinese Task: Kunshan Population Time Series}
\label{app:task_example_zh}
\begin{CJK*}{UTF8}{gbsn}
\begin{prompttemplate}{Task}
\normalfont\fontsize{9}{11.5}\selectfont
\textbf{Domain:} Policy and government data\par\medskip
我们在做昆山片区公共服务设施和住房需求测算，我需要把昆山市近5年（21-25年）的人口底数整理成一张可复核的时间序列。最好从 \url{http://www.ks.gov.cn/} 或其它政府网站上的公开数据入手，比如统计年鉴、统计公报，确认哪些年份可用、数据是网页还是下载文件、常住人口、户籍人口、流动人口等口径怎么写；口径对上以后，再提取昆山市历年的总人口、年龄结构、城乡结构和流动结构，并把各区镇分年度的同类指标汇总出来，看看哪些区镇变化更明显。遇到缺失年份、统计定义调整或区镇口径合并拆分时，我希望能标出来；如果某些细项拿不到，就说明限制并给出可替代的近似口径。
\end{prompttemplate}
\smallskip
\noindent\textbf{Selected rubrics.}
\par\smallskip
\begingroup
\fontsize{9}{11.5}\selectfont
\setlength{\tabcolsep}{5pt}
\renewcommand{\arraystretch}{1.12}
\noindent\begin{tabular}{@{}p{0.055\linewidth}>{\raggedright\arraybackslash}p{0.66\linewidth}>{\raggedright\arraybackslash}p{\dimexpr0.285\linewidth-20pt\relax}@{}}
\toprule
\textbf{ID} & \textbf{Rubric} & \textbf{Attributes} \\
\midrule
R7 & 回复应包含常住人口的统计口径是“在本市居住半年以上，或户口在本地且未外出半年以上” & \textit{Type:} Objective\par \textit{Necessity:} Explicit\par \textit{Priority:} Must Have \\
\addlinespace[6pt]
R25 & 回复应标明各数据的来源（如：xx年统计年鉴/xx年统计公报/第七次人口普查），给出可复核的数据来源路径url & \textit{Type:} Objective\par \textit{Necessity:} Implicit\par \textit{Priority:} Nice to Have \\
\bottomrule
\end{tabular}
\endgroup
\end{CJK*}

\subsection{English Task: University Tuition and Admissions Comparison}
\label{app:task_example_en}
\begin{prompttemplate}{Task}
\normalfont\fontsize{9}{11.5}\selectfont
\textbf{Domain:} Everyday facts\par\medskip
With application spreadsheets due this week in July 2026, I'm comparing my University of Oregon and UCF , and want the key figures aligned for each school: 2024--2025 out-of-state listed undergraduate tuition and admitted-student GPA/SAT results, including the SAT middle-50 range and midpoint. Using official university/NCES data with CollegeRaptor/PrepScholar only as backups, flag conflicts, gaps, and non-standard ranges.
\end{prompttemplate}
\smallskip
\noindent\textbf{Selected rubrics.}
\par\smallskip
\begingroup
\fontsize{9}{11.5}\selectfont
\setlength{\tabcolsep}{5pt}
\renewcommand{\arraystretch}{1.12}
\noindent\begin{tabular}{@{}p{0.055\linewidth}>{\raggedright\arraybackslash}p{0.66\linewidth}>{\raggedright\arraybackslash}p{\dimexpr0.285\linewidth-20pt\relax}@{}}
\toprule
\textbf{ID} & \textbf{Rubric} & \textbf{Attributes} \\
\midrule
R1 & The response should prioritize official university data and NCES/IPEDS official data; use backup data from CollegeRaptor and PrepScholar and mark clearly with "backup" when official data is missing. & \textit{Type:} Objective\par \textit{Necessity:} Explicit\par \textit{Priority:} Must Have \\
\addlinespace[6pt]
R4 & The response should calculate the SAT midpoint as the arithmetic mean of the 25th and 75th percentiles, and shall not directly use reported medians, and the calculation rule shall be clearly stated. & \textit{Type:} Objective\par \textit{Necessity:} Implicit\par \textit{Priority:} Must Have \\
\addlinespace[6pt]
R9 & The response should present institutional data in a tabular format with one row per school, for aligned comparison of key indicators. & \textit{Type:} Objective\par \textit{Necessity:} Implicit\par \textit{Priority:} Nice to Have \\

\bottomrule
\end{tabular}
\endgroup

\end{document}